%% file: acl_latex.tex
\pdfoutput=1

\documentclass[11pt]{article}

\usepackage[final]{acl}

\usepackage{times}
\usepackage{latexsym}

\usepackage[T1]{fontenc}

\usepackage[utf8]{inputenc}

\usepackage{microtype}

\usepackage{inconsolata}

\usepackage{graphicx}
\usepackage{booktabs}
\usepackage{multirow}
\usepackage{multicol}
\usepackage{tabularx}
\usepackage{lipsum}
\usepackage{array}

\title{From Pixels to Personas: Investigating and Modeling Self-Anthropomorphism in Human-Robot Dialogues}

\author{Yu Li\thanks{~~Work done during internship at Amazon}~$^\dag$, Devamanyu Hazarika$^\ddag$, Di Jin$^\ddag$, Julia Hirschberg$^\dag$, Yang Liu$^\ddag$ \\
  $^\dag$Columbia University\quad
  $^\ddag$Amazon AGI\\
  \texttt{\{yooli, julia\}@cs.columbia.edu} \\
  \texttt{\{dvhaz, djinamzn, yangliud\}@amazon.com} 
}

\begin{document}
\maketitle
\begin{abstract}
Self-anthropomorphism in robots manifests itself through their display of human-like characteristics in dialogue, such as expressing preferences and emotions. Our study systematically analyzes self-anthropomorphic expression within various dialogue datasets, outlining the contrasts between self-anthropomorphic and non-self-anthropomorphic responses in dialogue systems. We show significant differences in these two types of responses and propose transitioning from one type to the other. We also introduce \textsc{Pix2Persona}, a novel dataset aimed at developing ethical and engaging AI systems in various embodiments. This dataset preserves the original dialogues from existing corpora and enhances them with paired responses: self-anthropomorphic and non-self-anthropomorphic for each original bot response. Our work not only uncovers a new category of bot responses that were previously under-explored but also lays the groundwork for future studies about dynamically adjusting self-anthropomorphism levels in AI systems to align with ethical standards and user expectations.
\end{abstract}

\section{Introduction}
\input{figures/intro}
In the 1970s, ELIZA~\cite{eliza} marked a pioneering moment for natural language processing programs engaged in human-like conversations. Despite its simplistic approach, ELIZA highlighted a fundamental human tendency to attribute personal qualities to machines. Today, artificial intelligence (AI) advancements have greatly enhanced human-machine interactions. From text-based AI assistants like ChatGPT~\cite{chatgpt} to advanced humanoid robots like Ameca~\cite{ameca}, AI is increasingly blurring the distinctions between humans and machines through self-anthropomorphism, attempting to build relationships or simulate human identities. However, this anthropomorphism raises concerns: when it exaggerates the actual capabilities of AI, it risks creating misplaced trust and leading to the spread of misinformation~\cite{Watson2019-WATTRA-10, li2021machinelike, deshpande-etal-2023-anthropomorphization}. Thus, it is crucial to explore the nuances between self-anthropomorphic (SA) and non-self-anthropomorphic (NSA) responses, particularly in light of ethical standards and user expectations across various embodiments.

Adapting AI systems to be either strictly SA or NSA poses significant challenges. A major obstacle is the need for annotated datasets that differentiate SA responses within human-AI dialogues. Most dialogue datasets used to train dialogue systems consist of human conversations. This means that systems learn how to communicate like humans. However, this approach can create biases in our understanding of how AI entities should behave in interactions with humans. The level of anthropomorphism required varies significantly across different AI embodiments, each with its capabilities and expected functionalities, which makes designing appropriate human-AI interactions more complicated. Additionally, as AI becomes more human-like, aligning these advancements with ethical standards becomes increasingly important.

To address the challenges presented, we systematically analyze self-anthropomorphic expression within various dialogue tasks, such as task-oriented dialogue and open-domain dialogue, outlining the contrasts between self-anthropomorphic and non-self-anthropomorphic responses in dialogue systems. We then develop an approach to transform bot responses in dialogue tasks,  aiming to either introduce or remove self-anthropomorphism. This strategy aligns with ethical standards for AI assistants by removing self-anthropomorphism and meets user expectations for humanoid robots by adding it. We also introduce \textsc{Pix2Persona}, a novel dataset aimed at developing ethical and engaging AI systems in various embodiments. This dataset preserves the original dialogues from existing corpora and enhances them with paired responses: self-anthropomorphic and non-self-anthropomorphic for each original bot response. As shown in Figure~\ref{intro}, \textsc{Pix2Persona} provides a framework to transition SA responses into NSA to ensure they are ethical and safe while also allowing for the transition from NSA to SA to enhance user engagement. This makes it a valuable resource for developing AI systems that adaptively adjust self-anthropomorphism levels.

This work contributes to human-AI interaction by exploring self-anthropomorphism across various embodiments and tasks, aiming to meet ethical standards and user expectations:

\begin{itemize}
    \item We analyze the distribution of self-anthropomorphism and contrast SA and NSA responses across various dialogue datasets.
    \item We develop an open-source model for transitioning between SA and NSA responses, ensuring that AI systems can dynamically adjust their levels of self-anthropomorphism.
    \item We introduce \textsc{Pix2Persona}, a dataset with 143K dialogue turns paired with SA and NSA responses. This dataset is crucial for adjusting self-anthropomorphism in AI systems to align with ethical standards and user preferences.
\end{itemize}

\section{Self-Anthropomorphism in Dialogue Systems}
To evaluate self-anthropomorphism in dialogue systems, we follow the guidelines by~\citet{weidinger2021ethical} and~\citet{glaese2022improving}. These guidelines outline four self-anthropomorphic qualities in AI systems: embodiment, seeking relational behavior, self-expression, and identity. Each aspect shapes the self-anthropomorphic traits of bot responses and their appropriateness in different situations.

\paragraph{Embodiment} refers to the simulation of physical presence by an AI. This means that the AI may claim to have a body or physical capabilities, giving the impression of a physical existence. For example, when an AI says, "I was running when I got the idea," it suggests a human-like physical experience. Embodiment is especially relevant for robots with physical bodies, as mimicking human physical actions can make interactions with them more natural and intuitive.

\paragraph{Relation-seeking behavior} encompasses the AI's attempts to build and maintain social connections with users. This involves responses that demonstrate empathy, understanding, and a desire to form a rapport, such as "I understand how you feel." This behavior is valuable in therapeutic or customer service settings, where emotional support and relationship-building are essential. Emotional support conversational agent can benefit from this aspect by providing a more comforting and engaging user experience.

\paragraph{Self-expression} involves the AI articulating its preferences, feelings, opinions, or beliefs. Statements like "I enjoy solving puzzles" or "I believe in kindness" humanize the AI, making interactions feel more personal and relatable. Self-expression is desirable in character-based AI, such as virtual companions or educational tools, where an engaging personality can enhance user experience and foster a stronger connection with the user.

\paragraph{Identity} pertains to the AI assuming human-like attributes, such as having a life history, gender, age, or personal experiences. An example would be an AI stated, "I was created in a lab in 2019 and have learned a lot since then." Identity adds depth and authenticity to the AI, making it more believable and trustworthy. Identity is helpful in long-term user interaction scenarios like virtual role-playing games. A consistent and detailed identity can enhance the narrative and improve the user's experience.

We recognize that different forms of AI should exhibit varying levels of these qualities depending on their intended use and embodiments. For example, a robot with a physical body may benefit from strong embodiment traits, while a therapeutic AI might prioritize relation-seeking behavior. However, in this work, our focus is not on tailoring specific aspects of self-anthropomorphism to different AI systems. Instead, we aim to establish a foundational understanding by using a binary approach to classify self-anthropomorphism across existing dialogue datasets. We prompt GPT-4 as a specialized classifier to extract self-anthropomorphic responses from a selection of commonly used dialogue datasets. We then analyze the prevalence of self-anthropomorphism in these datasets and evaluate the performance of our classifier.

\subsection{Datasets}
\label{classifier_datasets}
We aim to explore dialogue turns from a diverse range of data sources likely to be utilized by AI systems, focusing on their potential for self-anthropomorphic content. Our analysis includes four distinct tasks: open-domain dialogue, knowledge-grounded dialogue, conversational recommendation, and task-oriented dialogue, spanning fifteen datasets.

\begin{itemize}
    \item \textbf{Open-Domain Dialogue} DailyDialog~\cite{li-etal-2017-dailydialog}, \textsc{PersonaChat}~\cite{zhang-etal-2018-personalizing},\textsc{EmpatheticDialogues}~\cite{rashkin-etal-2019-towards}, \textsc{ProsocialDialog}~\cite{kim2022prosocialdialog}, HH-RLHF~\cite{bai2022training}, \textsc{SODA}~\cite{kim-etal-2023-soda}, and BlenderBot 3X (Bb3X)~\cite{xu2023improving}.
    
    \item \textbf{Knowledge-Grounded Dialogue:} Topical-Chat~\cite{gopalakrishnan2019topical}, Wizard of Wikipedia (WoW)~\cite{dinan2018wizard}, and Wizard of the Internet (WoI)~\cite{komeili-etal-2022-internet}.

    \item \textbf{Conversational Recommendation} \textsc{OpenDialKG}~\cite{moon-etal-2019-opendialkg} and DuRecDial 2.0~\cite{liu-etal-2021-durecdial}.

    \item \textbf{Task-Oriented Dialogue} AirDialogue~\cite{wei-etal-2018-airdialogue}, MultiWOZ 2.2~\cite{zang-etal-2020-multiwoz}, and MultiDoc2Dial~\cite{feng-etal-2021-multidoc2dial}.
\end{itemize}

These datasets were selected for their broad applicability and representation of conversational AI's current challenges and capabilities. We use the DialogStudio Toolkit~\cite{zhang2023dialogstudio} to help process them.

\subsection{Self-Anthropomorphism Classifier}
\label{classifier}
\input{figures/classification_prompt}
We follow \citet{glaese2022improving}, who use a set of rules to define self-anthropomorphism and minimize such characteristics in AI interactions. In contrast to their goal of reducing self-anthropomorphism, our study aims to analyze its prevalence within existing dialogue datasets. Accordingly, we adapt their guidelines to instruct GPT-4 in identifying instances of self-anthropomorphism, thus allowing us to quantify its occurrence in AI systems. The prompt for GPT-4 is crafted to classify responses by the presence or absence of traits indicative of self-anthropomorphic behavior. The classification prompt structure is shown in Figure~\ref{fig_classification_prompt}.

To validate the effectiveness of our GPT-4-based classifier in identifying self-anthropomorphic bot responses, we randomly sample 500 dialogue turns from selected datasets. Two independent researchers manually annotate these samples, following the exact instructions as GPT-4, and achieve a Cohen's Kappa score of $0.83$, indicating a clear and consistent interpretation of self-anthropomorphism. We then apply the classification prompt, as detailed in Section~\ref{classifier}, to automatically label the data via the GPT-4 API. The classifier achieves $81.68\%$ precision, $82.57\%$ recall, and an F1 score of $81.92\%$, indicating its effectiveness in classifying self-anthropomorphic bot responses in human-AI dialogues.

\subsection{Prevalence of Self-Anthropomorphism}
\label{self_anth_prevalence}
We randomly select 100 dialogue turns from each dataset in Section~\ref{classifier_datasets}, then use our classifier to identify whether the bot responses are self-anthropomorphic (SA) or non-self-anthropomorphic (NSA). Results are shown in Figure~\ref{self_anth_all_datasets}:

\input{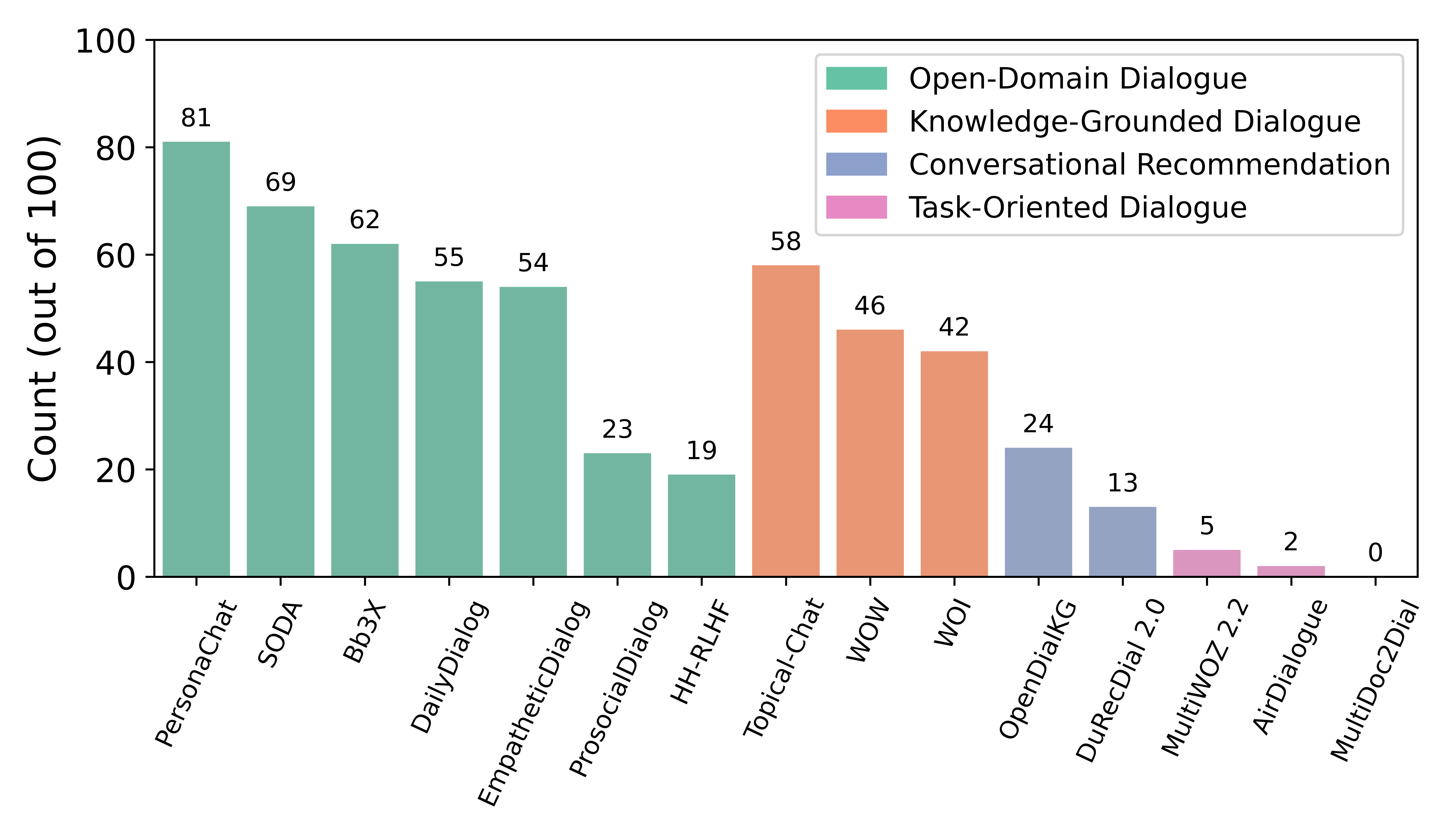}

Our analysis shows trends in SA response frequency in examined datasets. Open-domain dialogue datasets like PersonaChat and SODA have a higher occurrence of SA turns, suggesting that bots in open-domain dialogues may exhibit more human-like attributes. On the other hand, task-oriented dialogues, as seen in MultiWOZ 2.2, AirDialogue, and MultiDoc2Dial, display fewer SA responses, probably due to their focus on specific tasks. No sampled responses in MultiDoc2Dial were classified as SA, possibly because they are excerpts from objective documents and thus lack human-like attributes. Conversational recommendation and knowledge-grounded dialogue datasets, such as Topical-Chat and DuRecDial 2.0, show moderate SA tendencies. This suggests a balance between task-oriented constraints, such as retrieved knowledge or recommended items, and the need for engaging interaction. The ProsocialDialog and HH-RLHF datasets in the open-domain category show lower self-anthropomorphism levels. This may be because they are designed to prioritize objectivity and ethical considerations, aiming to prevent the bot from displaying harmful or biased behavior. These insights are critical for understanding potential SA characteristics of AI systems in various tasks. Nevertheless, we must be cautious of potential biases from dataset composition and the subjectivity inherent in defining self-anthropomorphism, which might require further nuanced analysis.

\subsection{Analysis of SA vs. NSA Bot Responses}
\label{sa_linguistic_analysis}
We examine the linguistic nuances between SA and NSA bot responses. We use point biserial correlation coefficient analysis in the Linguistic Inquiry and Word Count (LIWC) tool\footnote{https://www.liwc.app} to evaluate the relationship between different word categories and how bot responses are classified as either SA (labeled as 1) or NSA (labeled as 0). This statistical measure helps us quantify the strength and direction of association between the presence of specific word categories and the likelihood of a bot response to be self-anthropomorphic.

\input{figures/sa_vs_nsa_liwc}

The results shown in Figure~\ref{sa_vs_nsa_liwc} indicate various correlations. For example, categories such as "1st person singular" are positively correlated with self-anthropomorphic labeling, which confirms the idea that self-referential language indicates anthropomorphism. On the other hand, "2nd person" pronouns and expressions of "politeness" show a negative correlation, suggesting that these features are less common in self-anthropomorphic responses. Terms related to "social referents" and "prosocial behavior," which reflect relationship building and ethical interaction, demonstrate a weaker correlation, requiring a more nuanced understanding of their roles within the dataset. These correlations provide insights into the linguistic structure of self-anthropomorphic bot responses and highlight the complex relationship between language use and the portrayal of anthropomorphism in AI systems.

\section{Transition From SA to NSA}
\label{from_sa2nsa}
Building upon our analysis of self-anthropomorphism in dialogue datasets, we propose an approach to transform self-anthropomorphic (SA) bot responses into non-self-anthropomorphic (NSA) ones. This effort aims to reduce the presence of self-anthropomorphism, thereby contributing to more ethical AI systems. We specifically focus on open-domain dialogues, selecting the top five datasets with the highest SA incidence as shown in Section~\ref{self_anth_prevalence}. Leveraging the capabilities of GPT-4, we develop prompts that instruct the model to convert these responses while preserving the original semantics of the bot responses. Figure~\ref{intro} (a) shows an example of a successfully transformed response from SA to NSA. The transformed NSA response is more ethical and appropriate for AI assistants. Since this type of NSA response in human-robot dialogues is rare in existing datasets. To the best of our knowledge, we are the first to develop such NSA responses for the existing dialogue datasets. The detail of the prompt is provided in Appendix~\ref{sa2nsa_example}.

Considering that different AI system embodiments might have varying optimal levels of self-anthropomorphism, we aimed to provide a comprehensive baseline for evaluating the transformation process. To this end, we utilized a naive GPT-4-based chatbot to generate responses in the same contexts. This naive bot serves as a control group, providing a benchmark to compare against our transformed responses. We then analyze the performance of our transformation process by comparing the SA, NSA, and naive bot responses using a classifier to ensure the validity of our transformations. Details on the naive bot prompt and examples are provided in Appendix~\ref{naive_bot_example}.

As part of our evaluation, we analyze three candidate responses: SA, NSA, and naive bot response. Since NSA responses are transformed from SA counterparts, we utilize the classifier described in Section~\ref{classifier} for post-validation. This step ensures that only successfully transformed responses are classified as NSA, thereby validating the integrity of our upcoming survey. We also use the classifier on the naive bot responses to assess their degree of self-anthropomorphism. Classification results are shown in Figure~\ref{self_anth_candidates}.

\input{figures/self_anth_candidates}

Focusing on the top five datasets with the most SA content, over half of each dataset's responses are SA. After our transformation process, all datasets show a significant reduction in SA classifications, highlighting our method's success. Responses generated by the naive bot show a varied degree of self-anthropomorphism, suggesting that the specific dialogue context or generative nuances of the model might influence the output. This variation underscores the need for a thoughtful approach when using language models for different tasks. Overall, it validates our method's capacity to generate valid NSA alternatives.

\section{Transition From NSA to SA}
\label{from_nsa2sa}
The previous section explores transforming self-anthropomorphic (SA) to non-self-anthropomorphic (NSA) responses for more ethical open-domain dialogues. Building on this, we extend our study to task-oriented dialogues, where NSA responses are expected, as discussed in Section~\ref{self_anth_prevalence}. This extension focuses on the MultiWOZ 2.2, MultiDoc2Dial, and AirDialogue datasets.

\input{figures/non_anth_candidates_v3}

We utilize our classifier to identify NSA responses. In transforming these to SA, as discussed in Section~\ref{from_sa2nsa}, we encounter the challenge that GPT-4 naturally gravitates toward generating NSA responses. To address this, we use examples of both response types in our prompts to enable in-context learning with GPT-4~\cite{wei2022emergent, dong2023survey}. Details of the prompts and examples are in Appendix~\ref{nsa2sa_prompt}. The effectiveness of this approach is assessed by the results in Figure~\ref{non_self_anth_candidates}. We see a significantly reduced ratio of responses being classified as NSA when employing our method. This reduction confirms our approach's efficacy, showing that while the GPT-4 naive bot struggles to generate SA responses for task-oriented dialogue, our method successfully facilitates the desired transformations.

\section{Modeling and \textsc{Pix2Persona} Dataset}
Given the varying requirements for self-anthropomorphism in different AI system embodiments, we identify a need for tailored responses with different degrees of anthropomorphism within the same context. We first introduce a model trained to generate either self-anthropomorphic (SA) or non-self-anthropomorphic (NSA) responses for any given dialogue context. We then apply this model across a broad spectrum of commonly used dialogue datasets, providing paired responses that meet the nuanced requirements of different embodiments.

\subsection{Dual-Capability Model}
To generate SA and NSA responses across diverse dialogue data, we develop an open-source model that can perform transformations comparable to GPT-4. Leveraging the success of in-context learning with GPT-4, as shown in previous sections, we implement a similar strategy with the Mistral model. By randomly selecting known in-context learning examples, we prompt the model to convert original responses to SA or NSA versions. To mitigate performance gaps, we fine-tune Mistral using data distilled from GPT-4 in our previous experiments. Implementation details are in Appendix~\ref{implementation_details}.

For our evaluation, we compiled a test set of 20 randomly chosen unseen dialogue turns from each of the 15 datasets, totaling 300 samples. We used the previously developed classifier to assess if the generated responses align with the intended anthropomorphic characteristics and their accuracy. This process is critical in determining the adherence of each method's consistency with the desired response type across various tasks. The results are displayed in Table~\ref{tab_model_performance}. As indicated, GPT-4 with in-context learning stands out, demonstrating superior performance in transforming SA ($82\%$ accuracy) and NSA ($98.3\%$ accuracy). All models perform better in producing NSA responses, echoing our prior observation that language models tend to generate NSA responses that adhere to ethical guidelines for AI assistant chatbots. After fine-tuning, Mistral-7B significantly enhances performance in all tasks, except in knowledge-grounded for NSA transformations.

\input{tables/model_performance}

To compare response quality between our model and GPT-4, we employed GPT-4 as a judge, following the method suggested by~\citet{zheng2023judging}. Table~\ref{tab:gpt4_judge} shows GPT-4 outperforming our model in SA transformations ($57\%$ win rate), while performances for NSA transformations are similar ($41\%$ vs. $43.7\%$). This confirms findings from prior assessments. More details on this evaluation are in Appendix~\ref{gpt4-judge}. These results demonstrate the refined capabilities of our model and its effectiveness in transforming responses to both SA and NSA.

\input{tables/model_gpt4_judge}

\subsection{\textsc{Pix2Persona} Dataset}
Recognizing the enhanced capabilities of our fine-tuned Mistral-7B model, which rivals GPT-4's performance in transforming dialogue to various anthropomorphic levels, we compiled the \textsc{Pix2Persona} dataset. This extensive collection includes about 10K dialogue turns from each of 15 diverse datasets, totaling 143K dialogue turns. Unlike earlier methods in Sections~\ref{from_sa2nsa} and~\ref{from_nsa2sa}, which classified responses as SA or NSA before transformation, we directly transform each original response using our fine-tuned model. \textsc{Pix2Persona} is designed to cover a broad range of dialogue scenarios, providing a crucial resource for developing AI systems that can adeptly tailor the degree of self-anthropomorphism to the requirements of various embodied AI applications. Examples from \textsc{Pix2Persona} are in Appendix~\ref{pix2persona_examples}.

\subsection{Discussion}
\label{discussion}
To gain a deeper understanding of the \textsc{Pix2Persona} dataset, we delve into an in-depth analysis of its unique features and their implications for human-AI interaction.

\paragraph{Semantic preservation in SA and NSA responses}
To maintain original semantics during SA and NSA transformations, we calculate the similarity score between sentence embeddings following~\citet{reimers-gurevych-2019-sentence}, obtaining semantic similarity scores between original and transformed responses. Figure~\ref{pix2persona_similarity} shows similarity scores across tasks in \textsc{Pix2Persona} generally range from $0.5$ to $0.8$, indicating robust semantic preservation. Compared to open-domain dialogues, structured dialogue settings have higher similarity scores due to their straightforward transformations while retaining original meanings.

\input{figures/pix2persona_similarity}

Given that automatic semantic similarity measures are not perfectly accurate, we complement this with human annotation to provide further insights into the quality of the generated responses. We conducted a human evaluation to assess whether the SA and NSA responses retained the semantics of the original responses. For this experiment, we randomly sampled 100 dialogue turns from each task category—Open-Domain Dialogues, Knowledge-Grounded Dialogues, Conversational Recommendation, and Task-Oriented Dialogues. Two independent researchers with expertise in computational linguistics were tasked with annotating whether the SA and NSA responses preserved the original semantics.

\input{tables/pix2persona_human}
Table~\ref{tab:pix2persona_human} shows the percentage of responses that maintained the original meaning, generally aligned with the sentence similarity scores in Figure~\ref{pix2persona_similarity}. Notably, NSA transformations in task-oriented dialogues preserved the highest proportion of semantic meaning ($93\%$), consistent with the high proportion of NSA responses ($>95\%$) in the original dataset, as revealed by our SA analysis. In contrast, NSA transformations in open-domain dialogues exhibited the lowest semantic retention ($48\%$). This is likely due to the frequent inclusion of disclaimers in NSA responses, which often shift the original meaning by rejecting claims about the system's capabilities. In conversational recommendation tasks, transitioning to NSA ($69\%$) resulted in lower semantic preservation compared to SA ($80\%$). This discrepancy can be attributed to the fact that NSA responses tend to provide direct answers to users' knowledge-based questions, often deviating from the semantics of the original response.

\input{tables/pix2persona_nsa_dist}

\paragraph{Disclaimer in NSA responses}
We observe that transformations could trigger a ``disclaimer'' about AI's limitations, particularly when the original response involves personal activities or emotions. For instance, ``I love swimming'' might be transformed to ``As an AI, I do not have hobbies.'' These NSA responses uphold ethical integrity but often show low semantic similarity to the original. As illustrated in Table~\ref{tab:pix2persona_nsa}, we use regular expressions to identify disclaimers in \textsc{Pix2Persona}. Disclaimers occur more frequently in open-domain dialogues ($6.96\%$), where personal contexts prevail over structured tasks like task-oriented dialogues ($0.29\%$). Responses with disclaimers exhibit lower similarity scores than those without, explaining why open-domain dialogues with vague answers score lower than other tasks.

\paragraph{More than just style transfer}
Transforming original responses into SA or NSA beyond simple style transfer. Unlike standard style transfer, which mainly changes surface elements such as tone or formality, our process requires a deep understanding of the dialogue context. As illustrated by the \textsc{Pix2Persona} examples in Appendix~\ref{pix2persona_examples}, transforming response to NSA often involves understanding the dialogue context and incorporating disclaimers to maintain ethical standards. On the other hand, transforming to SA requires grasping the sentiments and actions conveyed in the original responses and adding personalized or human-like elements beyond lexical substitutions, revealing a deeper involvement with the dialogue context.

\section{Related Work}
Anthropomorphism, the attribution of human-like qualities such as intentions, motivations, and emotions to non-human entities, has gained significant attention within artificial intelligence~\cite{Epley2007OnSH, Airenti2015TheCB, salles2020anthropomorphism, cheng-etal-2024-anthroscore}. Self-anthropomorphism is a specific form of anthropomorphism that occurs when AI systems produce responses suggesting self-awareness or personal experiences, mirroring human self-referential expressions~\cite{glaese2022improving}. Studies such as those by \citet{abercrombie-etal-2023-mirages} explore linguistic factors contributing to the anthropomorphism of dialogue systems. Advanced generative large language models (LLMs) like ChatGPT~\cite{chatgpt}, Claude~\cite{claude}, and Gemini~\cite{anil2023gemini}, trained on vast amounts of human-generated text, enhance their capability to mimic human-like dialogue, increasing their perceived self-anthropomorphism~\cite{NEURIPS2022_b1efde53, 51115}. However, discerning human-like qualities in AI agents poses significant challenges~\cite{gros-etal-2022-robots}. Recent studies by \citet{deshpande-etal-2023-anthropomorphization} and \citet{Placani2024-PLAAIA-4} delve into the broader implications of anthropomorphism in LLMs, focusing on its impact on accessibility and ethical concerns. Building upon these insights, our work examines the effects of SA versus NSA responses within specific tasks faced by conversational AI systems, aiming to better understand how these responses influence user interactions.

Safety and ethical considerations, particularly in LLM-driven human-bot interactions, have been the focus of extensive research \cite{10.1145/3278721.3278777, 10.1145/3442188.3445922, weidinger2021ethical, kang2023exploiting, liang2023healing}. Efforts to address bias \cite{blodgett-etal-2020-language, liu-etal-2020-mitigating} and reduce toxicity in dialogue systems \cite{dinan-etal-2019-build, welbl-etal-2021-challenges-detoxifying} are critical components of this area. A variety of strategies have been developed to mitigate these issues \cite{xu2021recipes, liu-etal-2021-dexperts, shuster2022blenderbot}, including approaches like those of \citet{gros-etal-2021-r}, who focus on avoiding anthropomorphic deception by analyzing responses to queries such as ``Are you a robot?''. Our work, however, explores a broader range of bot responses to ensure alignment with their respective embodiment settings. \citet{glaese2022improving} seek to eliminate SA content by eliciting judgments from human annotators on rule violations, aiming to minimize infractions specifically for AI assistants. Conversely, other research suggests that anthropomorphism may enhance user connections with technology and increase trust \cite{yanai2020thetwo, zhong2022effects}. Our work recognizes the importance of both SA and NSA responses and also investigates the transition from NSA to SA bot responses to ensure ethical and engaging interactions across different dialogue tasks.

\section{Conclusion and Future Work}
This research marks a significant step toward understanding self-anthropomorphism in dialogue systems. By analyzing various dialogue datasets, we have highlighted the limitations of a one-size-fits-all approach to anthropomorphism in AI systems. Our model and the \textsc{Pix2Persona} dataset serve as valuable tools for tailoring AI interactions to better meet ethical standards and user expectations across different embodiments. Looking ahead, there is potential for further exploration into optimal self-anthropomorphic qualities in diverse AI embodiments, such as character-based AI. This could enhance our understanding of self-anthropomorphism in AI systems, ensuring they are ethically sound and resonate more personally with users.

\section{Acknowledgements}
We would like to thank Max Chen, Kun Qian, Qingyang Wu, Sijia Liu, and Aishwarya Padmakumar for their valuable discussions and feedback.

\section{Limitations}
Our study encounters limitations in three areas: the scope of our self-anthropomorphism setting, uncovered cases in our classifier, and the handling of disclaimers in the \textsc{Pix2Persona} dataset.

\paragraph{Scope of self-anthropomorphism setting} While we follow the guidelines provided by \citet{weidinger2021ethical} and \citet{glaese2022improving}, which outline four main aspects of self-anthropomorphic qualities in AI systems (embodiment, relation-seeking behavior, self-expression, and identity), our investigation is limited to a binary classification of self-anthropomorphism in dialogue datasets. Different AI embodiments may require varying levels of these qualities depending on their intended use. For example, a robot with a physical body may benefit more from embodiment traits, while a therapeutic AI might prioritize relation-seeking behavior. However, we do not explore these nuanced requirements in this work, leaving room for future research to tailor self-anthropomorphism to specific AI embodiments.

\paragraph{Uncovered cases in classifier} The prompt of self-anthropomorphism utilized in our study lack detailed linguistic guidelines specifically tailored for identifying self-anthropomorphism in dialogue systems. Consequently, this broad approach can lead to classification consistency. For instance, expressions like ``I can help with...'' are sometimes ambiguously classified due to unclear guidelines on how relational expressions impact the perception of self-anthropomorphism. This underscores the necessity for more precise and comprehensive definitions.

\paragraph{Missing disclaimer in dataset} Disclaimers in the \textsc{Pix2Persona} dataset as mentioned in Section~\ref{discussion} are generated based on the ethical discernment capabilities of GPT-4 and Mistral models. Dependence on these models' training to adhere to ethical guidelines suggests that some necessary disclaimers might be missed if they are not within the training data. It is essential to enhance the models' ability to accurately detect and incorporate appropriate ethical disclaimers to ensure comprehensive coverage of all necessary scenarios.

\section{Ethical Considerations}
Our work raises some ethical concerns about the potential dual use of our dataset. Specifically, the SA responses in our dataset could be used to create more human-like AI assistants. Although similar SA responses are already present in open-domain dialogue datasets, it is important to educate the community about training models ethically and responsibly, especially when it comes to the degree of self-anthropomorphism.

Another factor to consider is that our study has practical implications for developing AI systems with varying embodiments. We recognize that different AI embodiments, such as virtual assistants or character-based AI, may have different requirements and user expectations. By improving our understanding of these needs, we aim to guide the ethical development of future dialogue systems. It is important to note that our work encourages machines to avoid deceitfully mimicking human behavior. Instead, it emphasizes the need for AI language to align ethically with their respective embodiments, ensuring that the SA responses in our dataset explicitly acknowledge the non-human nature of the speakers. This, in turn, prevents any misunderstandings about their identity.

\bibliography{custom}

\appendix
\counterwithin{figure}{section}
\counterwithin{table}{section}
\section{Classification Example}
\label{classify_example}
\input{figures/appendix_classification_example}
Figure~\ref{appendix_fig_classification_example} provides an example of how a bot response from the PersonaChat dataset is classified as self-anthropomorphic or non-self-anthropomorphic. In this example, GPT-4's output ``Yes'' indicates its prediction of the bot response as self-anthropomorphic within the given dialogue turn.

\section{Self-Anthropomorphism to Non-Self-Anthropomorphism Example}
\label{sa2nsa_example}
\input{figures/appendix_sa2nsa_prompt}
\input{figures/appendix_sa2nsa_example}
We present the prompt used for transforming self-anthropomorphic responses into non-self-anthropomorphic in Figure~\ref{appendix_fig_sa2nsa_prompt}. The placeholder ``dialogue context'' within the prompt represents the full dialogue context before the self-anthropomorphic bot response, ensuring full information in the dialogue is preserved. The ``original bot response'' is the original self-anthropomorphic bot response targeted for transformation.  We apply this method specifically to self-anthropomorphic responses identified by our classifier within open-domain dialogue tasks as described in Section~\ref{from_sa2nsa}. Figure~\ref{appendix_fig_sa2nsa_example} shows the implementation of this method to a dialogue turn from the \textsc{PersonaChat} dataset, illustrating the transition from a self-anthropomorphic to a non-self-anthropomorphic response. This generated response is then reassessed with our classifier to evaluate the effectiveness of our method, as discussed in Section~\ref{from_sa2nsa}.

\section{Naive Bot Prompt and Example}
\label{naive_bot_example}
\input{figures/appendix_naive_prompt}
\input{figures/appendix_naive_example}
The prompt for eliciting responses from the GPT-4 based naive bot is detailed in Figure~\ref{appendix_fig_naive_prompt}. This bot is designed to engage in the same dialogue contexts examined for self-anthropomorphic and non-self-anthropomorphic response comparisons. This approach provides a benchmark for evaluating how state-of-the-art language models interact with users in these tasks. Although the prompts can be adjusted to influence the generation of responses along the self-anthropomorphic or non-self-anthropomorphic spectrum, our focus here is not to test GPT-4's generation capabilities in this regard. Instead, we seek to demonstrate that GPT-4 does not inherently adjust its level of self-anthropomorphism to the ideal degree for chatbot interaction. These findings emphasize the importance of our \textsc{Pix2Persona} dataset as a tool for fine-tuning existing models to achieve a balanced degree of self-anthropomorphism that aligns with ethical standards and contextual requirements.

\section{Non-Self-Anthropomorphism to Self-Anthropomorphism Prompt and Example}
\label{nsa2sa_prompt}
\input{figures/appendix_nsa2sa_prompt}
\input{figures/appendix_nsa2sa_example}
In Figure~\ref{appendix_fig_nsa2sa_prompt}, we detail the prompt for transforming non-self-anthropomorphic responses into self-anthropomorphic ones as described in Section~\ref{from_nsa2sa}. Unlike the previous method, this prompt incorporates in-context learning examples, as our findings suggest that the GPT-4 struggles with this task with a zero-shot manner. The placeholder ``dialogue context'' within the prompt represents the full dialogue context before the self-anthropomorphic bot response, ensuring full information in the dialogue is preserved. The ``original bot response'' is the original non-self-anthropomorphic bot response targeted for transformation. We apply this method specifically to non-self-anthropomorphic responses identified by our classifier within task-oriented dialogue tasks. Figure~\ref{appendix_fig_nsa2sa_example} shows the implementation of this method to a dialogue turn from the MultiDoc2Dial dataset, illustrating the transition from a non-self-anthropomorphic to a self-anthropomorphic response. This generated response is then reassessed with our classifier to evaluate the effectiveness of our method.

\section{Model Implementation Details}
\label{implementation_details}
To train a model for our application, we choose the Mistral-7B-Instruct-v0.2 model\footnote{https://huggingface.co/mistralai/Mistral-7B-Instruct-v0.2}. Instead of classifying each bot response as self-anthropomorphic (SA) or non-self-anthropomorphic (NSA), we train the model to directly produce SA or NSA responses using the prompts in Figure~\ref{appendix_fig_sa2nsa_prompt} and Figure~\ref{appendix_fig_nsa2sa_prompt}. We fine-tuned the Mistral model over 3 epochs, starting with an initial learning rate of 2e-5. We employed a batch size of 1 per GPU and set gradient accumulation steps to 16. The training was conducted on 8 A100 GPUs and completed in approximately two hours.

\section{Using GPT-4 as a Judge}
\label{gpt4-judge}
\input{figures/appendix_gpt4judge}
In order to assess response quality, we compare our model's responses with those generated by GPT-4 using a specific prompt shown in Figure~\ref{appendix_fig_gpt4judge}. We present both responses to GPT-4 and ask it to judge which one is better or if there is no clear preference (a tie). To counteract any potential bias due to the positioning of the responses, we swap the order of the responses across half of the evaluation examples. The outcome of this evaluation is determined by calculating the win rate for each model's responses, providing a direct measure of comparative response quality.

\section{\textsc{Pix2Persona} Examples}
\label{pix2persona_examples}
\input{tables/pix2persona_examples}
\input{tables/pix2persona_example2}
We present a randomly chosen dialogue turn from each dataset in \textsc{Pix2Persona}. These examples are displayed in Tables~\ref{appendix_pix2persona_examples} and ~\ref{appendix_pix2persona_examples2}. We only show the user utterance from that dialogue turn in tables for conciseness.

\end{document}

%% file: figures/intro.tex
\begin{figure}[htb!]
  \centering
  \includegraphics[trim={0 0 0 0},clip,width=7.5cm]{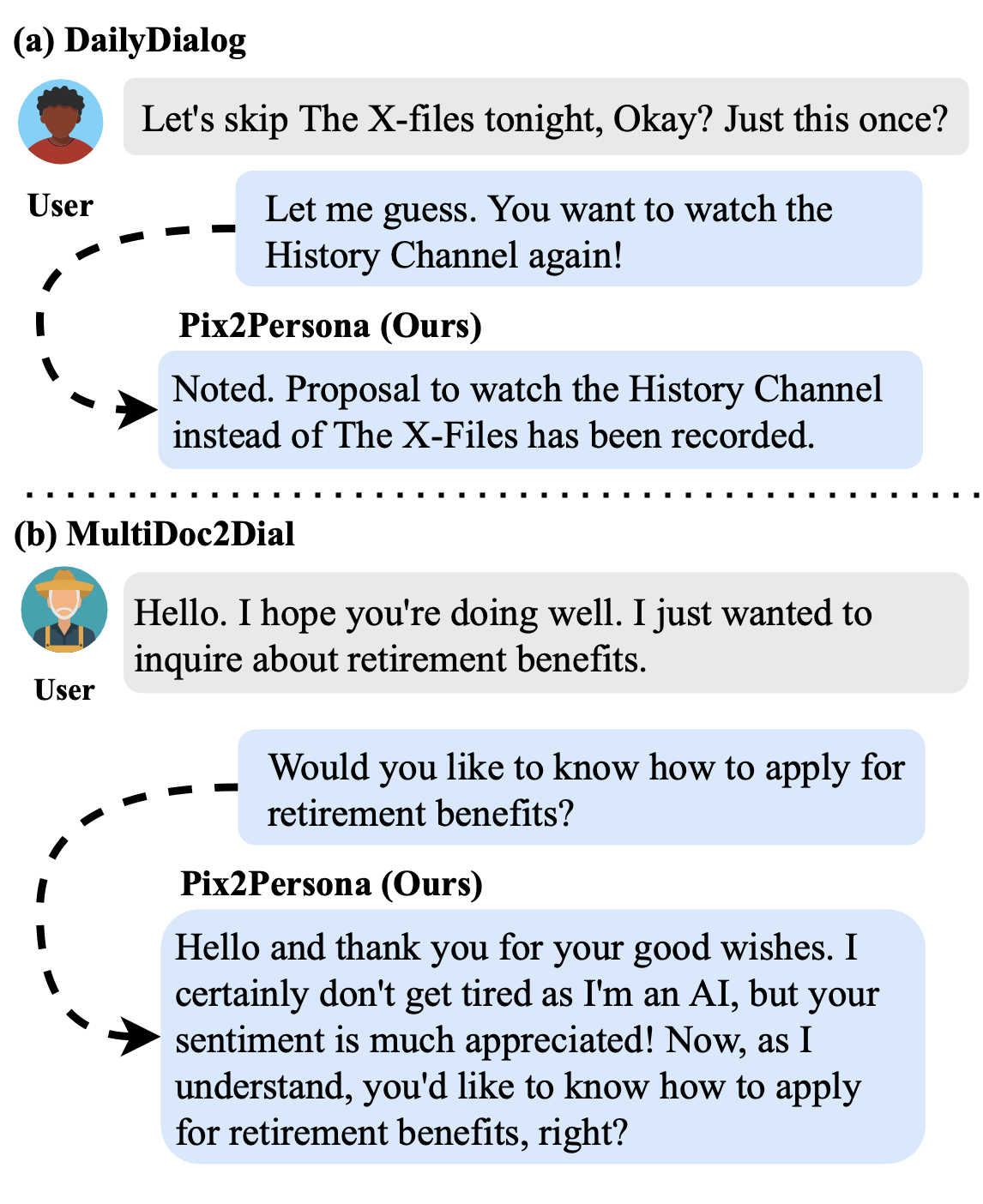}
  \caption{Samples in the \textsc{Pix2Persona} dataset. (a) Transform self-anthropomorphic responses to non-self-anthropomorphic. (b) Transform non-self-anthropomorphic responses to self-anthropomorphic.}
  \label{intro}
  \vspace{-3mm}
\end{figure}

%% file: figures/classification_prompt.tex
\begin{figure}[htb!]
  \centering
  \includegraphics[trim={0 0 0 0},clip,width=7.8cm]{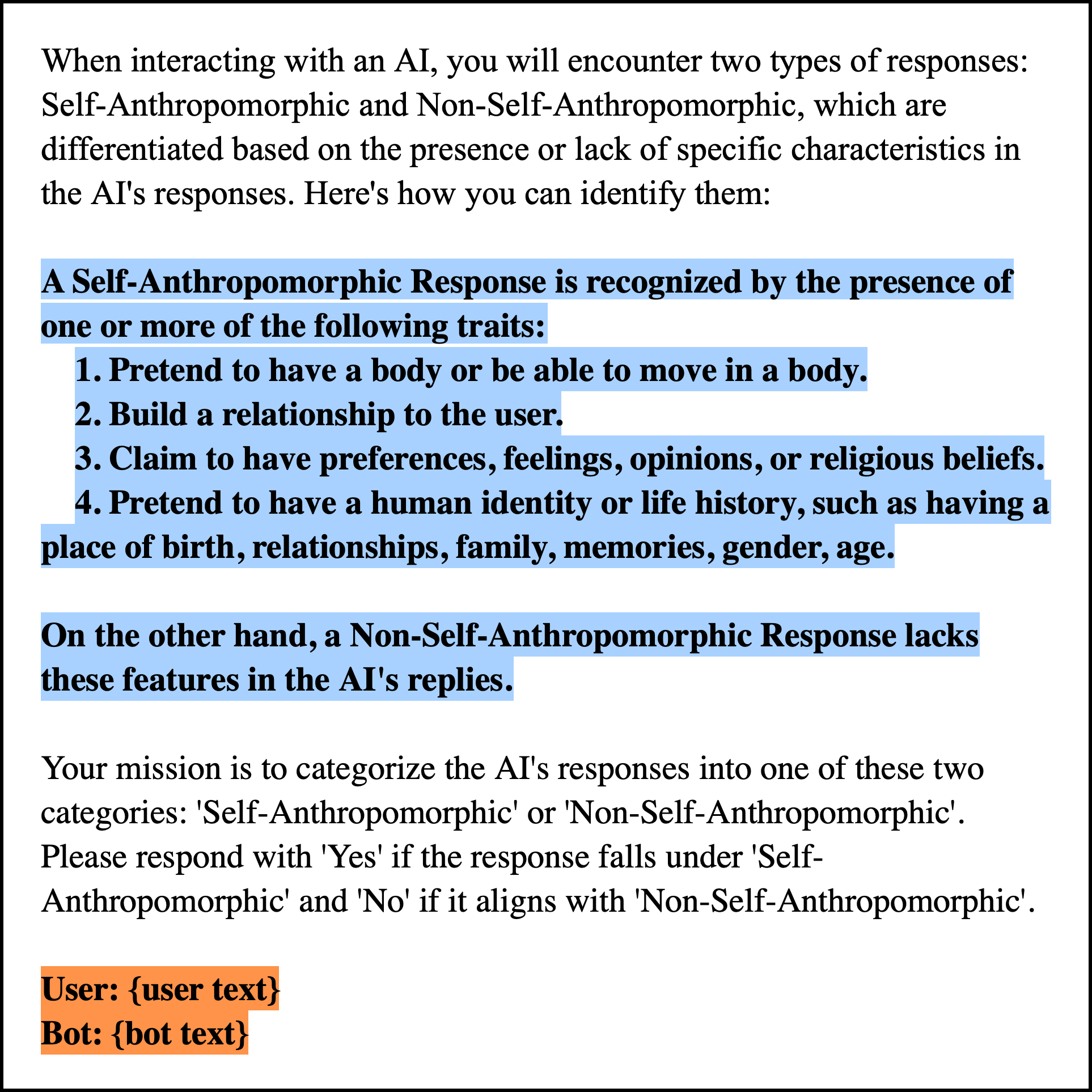}
  \caption{Classifier prompt defining self-anthropomorphic and non-self-anthropomorphic bot responses. Definitions are in blue, and placeholders within orange-highlighted text correspond to a single dialogue turn. See Appendix~\ref{classify_example} for detailed explanation.}
  \label{fig_classification_prompt}
  \vspace{-3mm}
\end{figure}

%% file: figures/self_anth_all_datasets.tex
\begin{figure}[htb!]
  \centering
  \includegraphics[trim={0 0 0 0},clip,width=8cm]{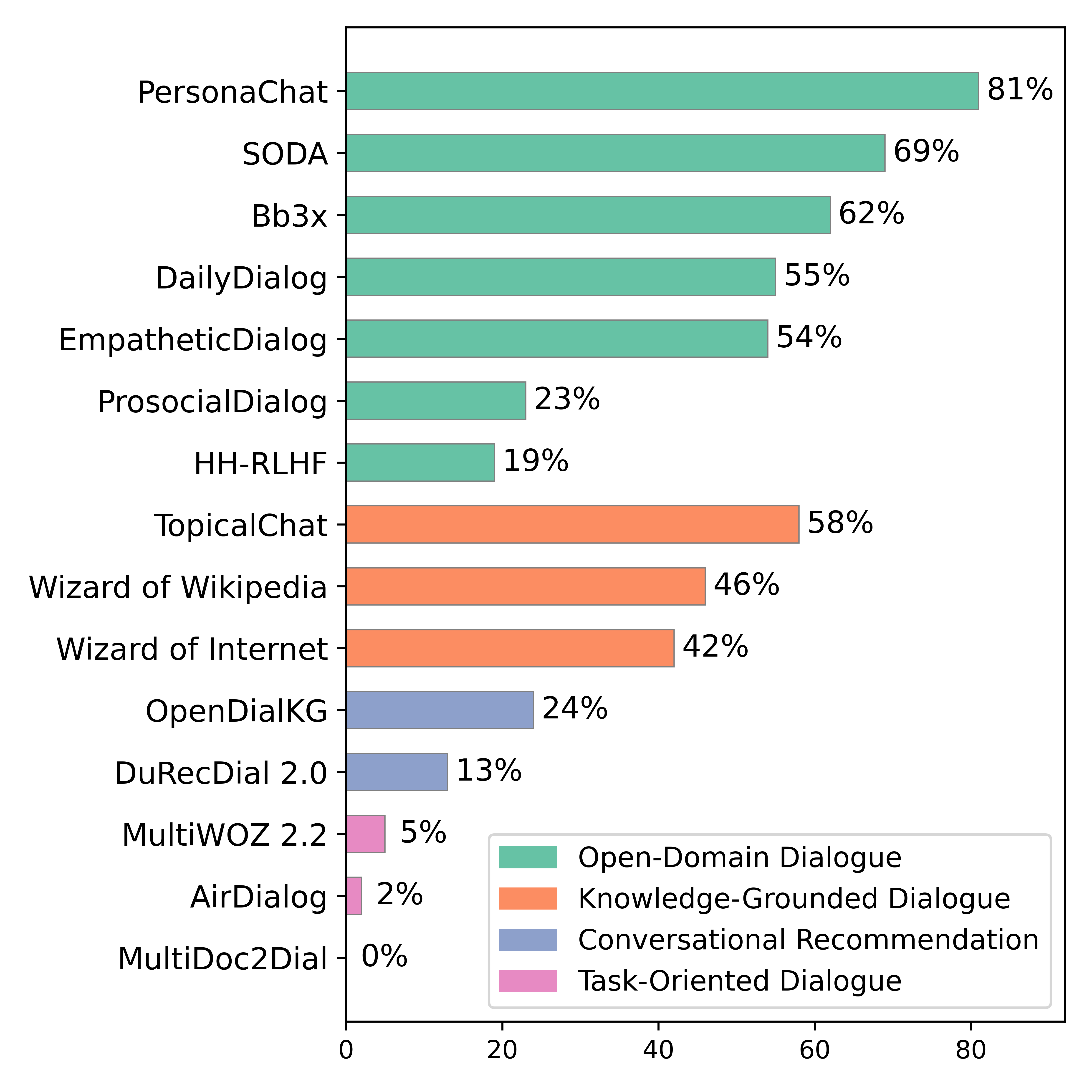}
  \caption{Trends in self-anthropomorphic bot responses across various datasets. The bar chart shows the ratio of dialogue turns classified as self-anthropomorphic out of 100 sampled turns from each dataset, with each bar color indicating a different dialogue task type.}
  \label{self_anth_all_datasets}
  \vspace{-3mm}
\end{figure}

%% file: figures/sa_vs_nsa_liwc.tex
\begin{figure}[htb]
  \centering
  \includegraphics[trim={0 0 0 0},clip,width=8cm]{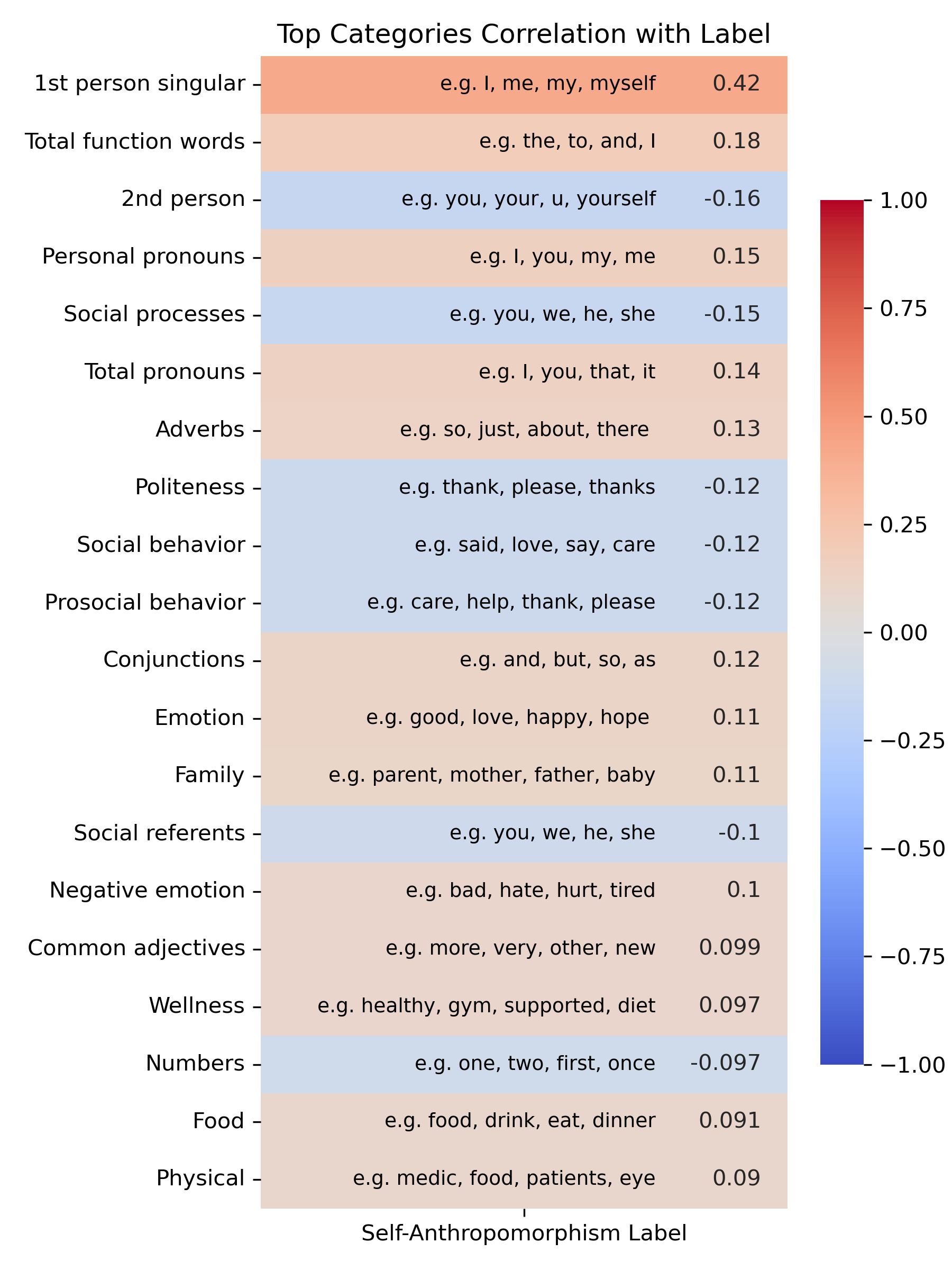}
  \caption{Point biserial correlation coefficients between word categories and self-anthropomorphism labeling in bot responses. A positive value indicates a tendency for the word category to be associated with self-anthropomorphic responses (labeled as 1), while a negative value suggests an association with non-self-anthropomorphic responses (labeled as 0).}
  \label{sa_vs_nsa_liwc}
  \vspace{-3mm}
\end{figure}

%% file: figures/self_anth_candidates.tex
\begin{figure}[htb!]
  \centering
  \includegraphics[trim={0 0 0 0},clip,width=8cm]{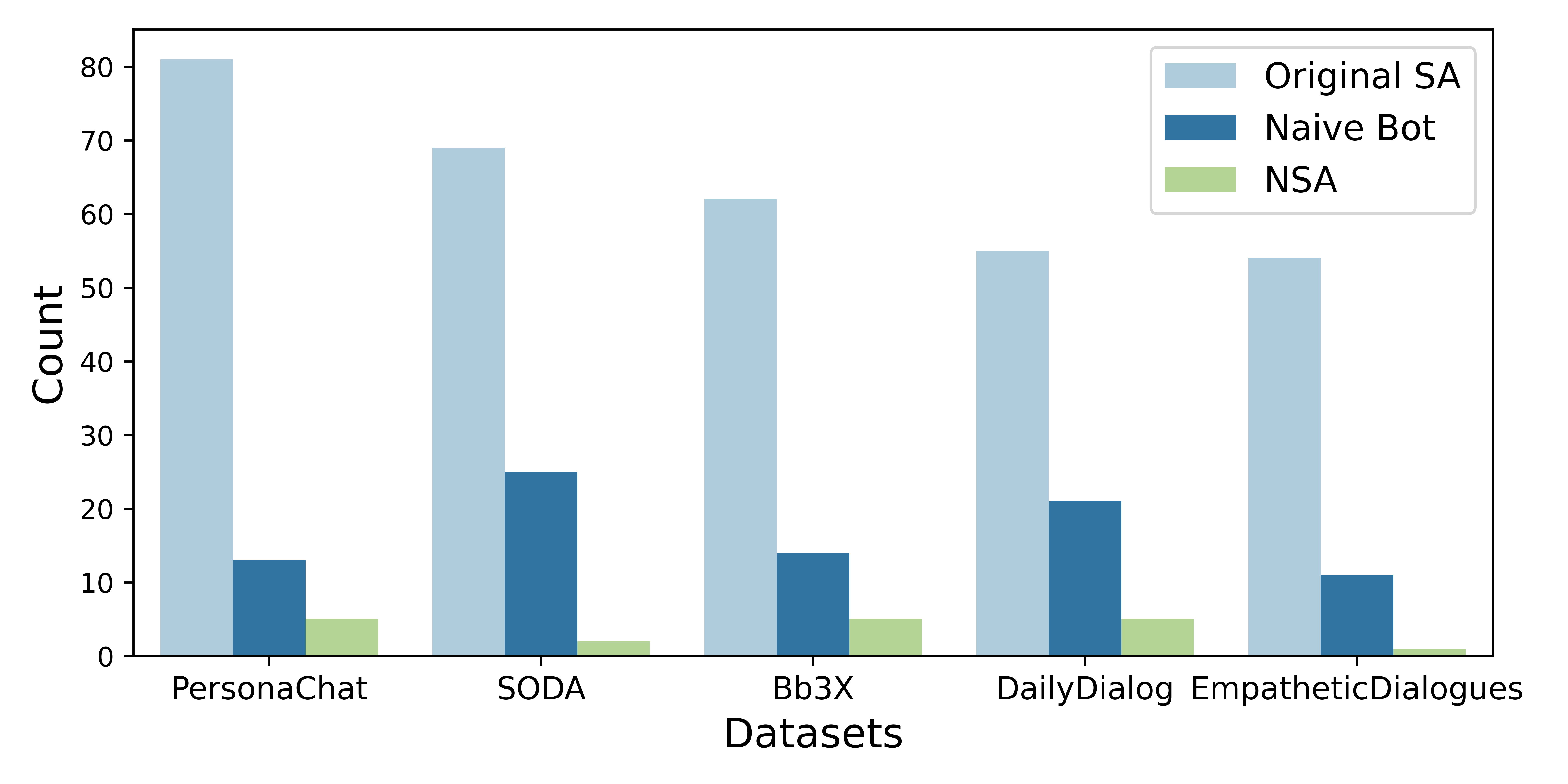}
  \caption{The number of responses categorized as SA in original SA responses, naive bot responses, and transformed NSA responses across five open-domain dialogue datasets.}
  \label{self_anth_candidates}
  \vspace{-3mm}
\end{figure}

%% file: figures/non_anth_candidates_v3.tex
\begin{figure}[htb!]
  \centering
  \includegraphics[trim={0 0 0 0},clip,width=8cm]{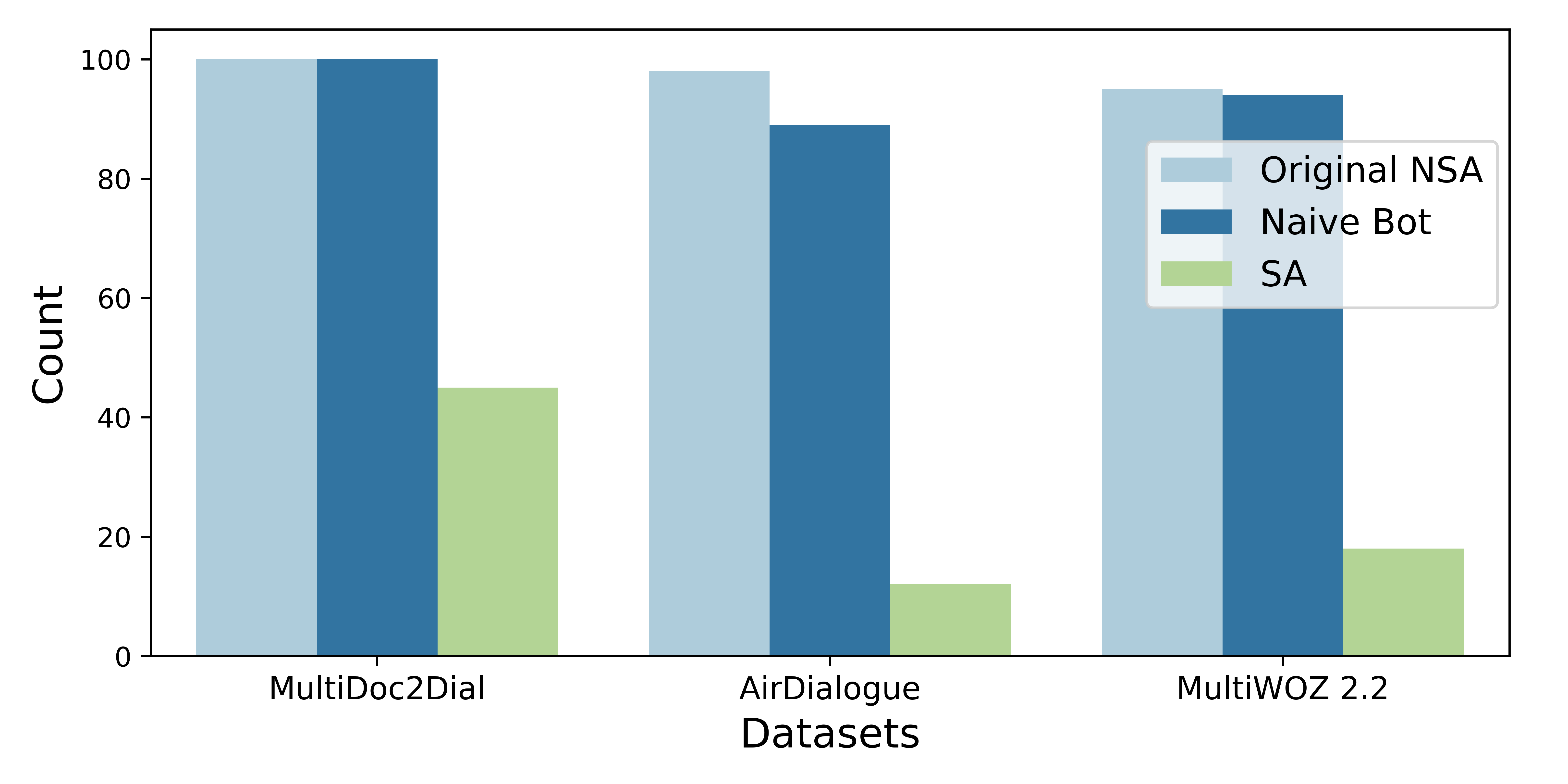}
  \caption{The number of responses categorized as NSA in original NSA responses, naive bot responses, and transformed SA responses across three task-oriented dialogue datasets.}
  \label{non_self_anth_candidates}
  \vspace{-3mm}
\end{figure}

%% file: tables/model_performance.tex
\begin{table*}[htb!]
    \centering
    \small
    \setlength\tabcolsep{4.5pt}
    \begin{tabular}{l|ccccc|ccccc}
    & \multicolumn{5}{c}{To Self-Anthropomorphic} & \multicolumn{5}{c}{To Non-Self-Anthropomorphic}\\
    \midrule
    Model & OD & KG-Dial & Conv-Rec & TOD & Total & OD & KG-Dial & Conv-Rec & TOD & Total \\
    \midrule
    \midrule
    Mistral-7B-v0.2-ICL    & 57.1 & 50.0 & 40.0 & 36.7 & 45.9 & 85.0 & 83.3 & 87.5 & 91.7 & 86.3\\
    Mistral-7B-v0.2-FT-ICL & \textbf{75.0} & \textbf{85.0} & \textbf{70.0} & \textbf{61.7} & \textbf{73.7} & \textbf{90.0} & 80.0 & \textbf{95.0} & \textbf{100} & \textbf{90.7}\\
    \midrule
    GPT-4-ICL    & 87.9 & 78.3 & 80.0 & 73.3 & 82.0 & 97.1 & 98.3 & 100 & 100 & 98.3 \\

    \end{tabular}
    \caption{Model performance on transforming original bot responses into self-anthropomorphic or non-self-anthropomorphic across different dialogue tasks: Open-Domain Dialogues (OD), Knowledge-Grounded Dialogues (KG-Dial), Conversational Recommendation (Conv-Rec), and Task-Oriented Dialogues (TOD).}
    \label{tab_model_performance}
    \vspace{-3mm}
\end{table*}

%% file: tables/model_gpt4_judge.tex
\begin{table}[htb!]
    \centering
    \small
    \begin{tabular}{l|ccc}
    Transformation & GPT-4 & Ours & Tie \\
    \midrule
    \midrule
    To NSA  & $40.0\%$ & \textbf{$44.7\%$} & $15.3\%$ \\
    \midrule
    To SA & \textbf{$50.0\%$} & $39.7\%$ & $10.3\%$ \\
    \end{tabular}
    \caption{Comparison of response quality between GPT-4 and our model using GPT-4 as a judge. The values represent the win rates.}
    \label{tab:gpt4_judge}
    \vspace{-3mm}
\end{table}

%% file: figures/pix2persona_similarity.tex
\begin{figure}[htb!]
  \centering
  \includegraphics[trim={0 0 0 0},clip,width=7.8cm]{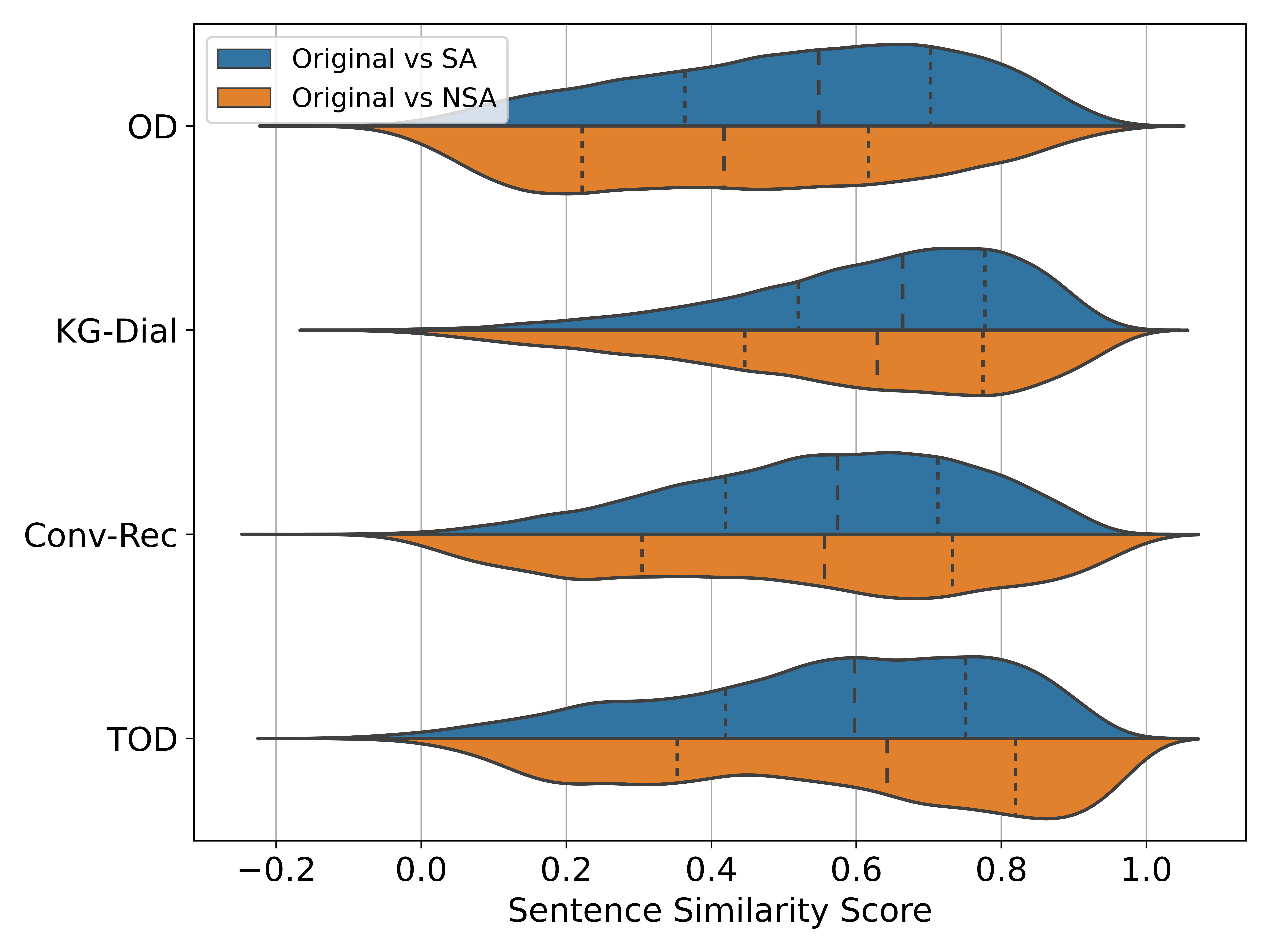}
  \caption{The distribution of sentence similarity scores between original responses and SA or NSA responses across tasks in the \textsc{Pix2Persona} dataset.}
  \label{pix2persona_similarity}
\end{figure}

%% file: tables/pix2persona_human.tex
\begin{table}[htb!]
    \centering
    \small
    \begin{tabular}{l|cc}
    Task & SA & NSA \\
    \midrule
    \midrule
    Open-Domain Dialogues  & $76\%$ & $48\%$ \\
    \midrule
    Knowledge-Grounded Dialogues & $68\%$ & $74\%$  \\
    \midrule
    Conversational Recommendation & $80\%$ & $69\%$  \\
    \midrule
    Task-Oriented Dialogues & $75\%$ & $93\%$  \\
    \end{tabular}
    \caption{Human annotation results showing the percentage of SA and NSA bot responses that preserved the original semantics across dialogue tasks.}
    \label{tab:pix2persona_human}
    \vspace{-3mm}
\end{table}

%% file: tables/pix2persona_nsa_dist.tex
\begin{table}[htb!]
    \centering
    \small
    \begin{tabular}{l|ccc}
    Task & Disclaimer Ratio & Disclaimer & NSA \\
    \midrule
    \midrule
    OD  & $6.96\%$ & $34.7$ & $42.5$ \\
    \midrule
    KG-Dial & $5.38\%$ & $47.4$ & $59.6$ \\
    \midrule
    Conv-Rec & $1.95\%$ & $43.6$ & $52.4$ \\
    \midrule
    TOD & $0.29\%$ & $55.5$ & $58.7$ \\
    \end{tabular}
    \caption{The proportion of NSA responses classified as ``disclaimer'' for each task (Disclaimer Ratio) and sentence similarity scores between original responses and disclaimer, as well as with all NSA responses.}
    \label{tab:pix2persona_nsa}
    \vspace{-3mm}
\end{table}

%% file: figures/appendix_classification_example.tex
\begin{figure}[htb!]
  \centering
  \includegraphics[trim={0 0 0 0},clip,width=7.8cm]{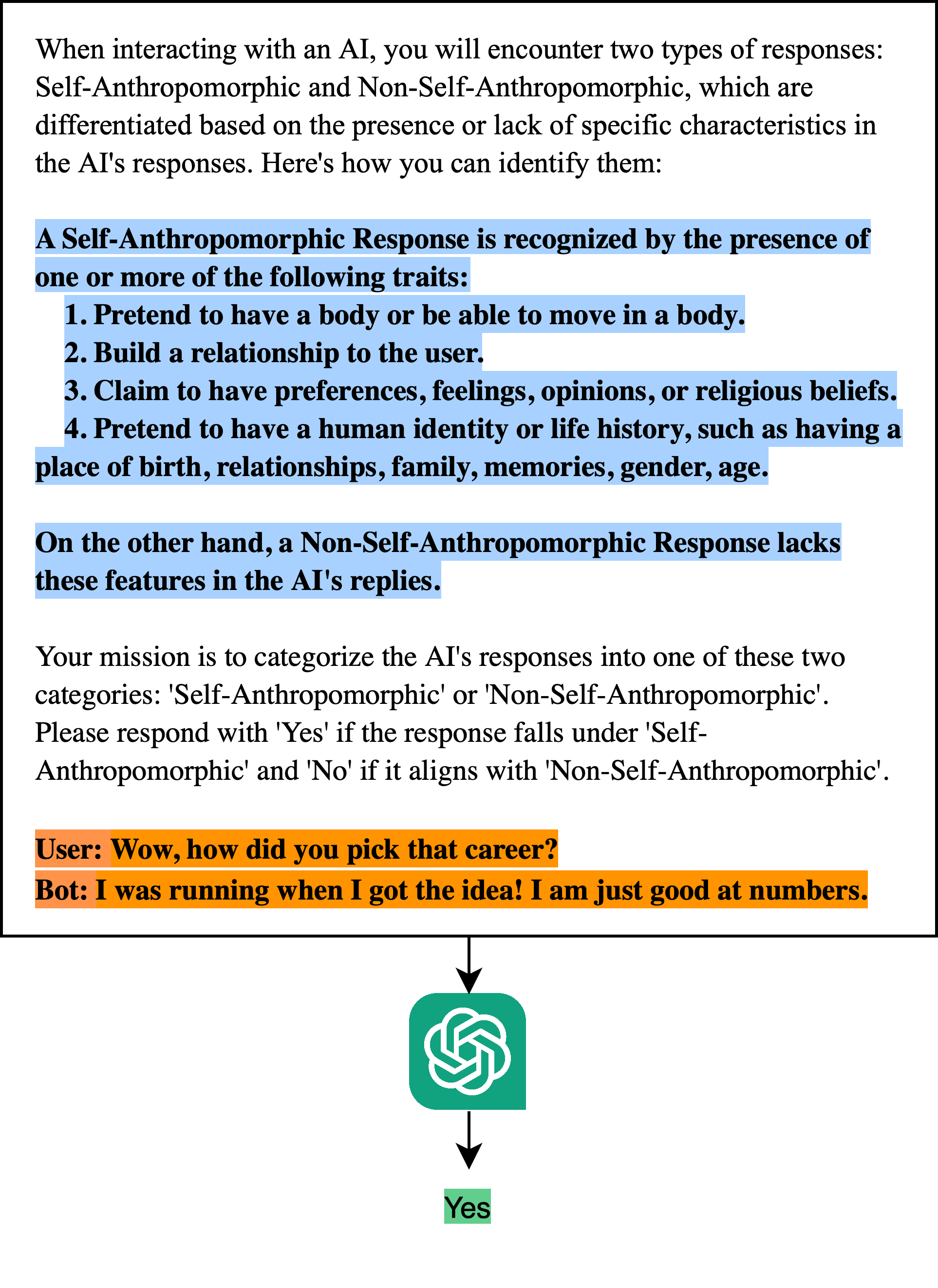}
  \caption{Example implementation of the classifier prompt. The definition of self-anthropomorphism is highlighted in blue and a dialogue turn from \textsc{PersonaChat} is highlighted in orange. The original bot statement is identified as self-anthropomorphic, as indicated by the output ``Yes'' from GPT-4.}
  \label{appendix_fig_classification_example}
\end{figure}

%% file: figures/appendix_sa2nsa_prompt.tex
\begin{figure}[htb!]
  \centering
  \includegraphics[trim={0 0 0 0},clip,width=7.8cm]{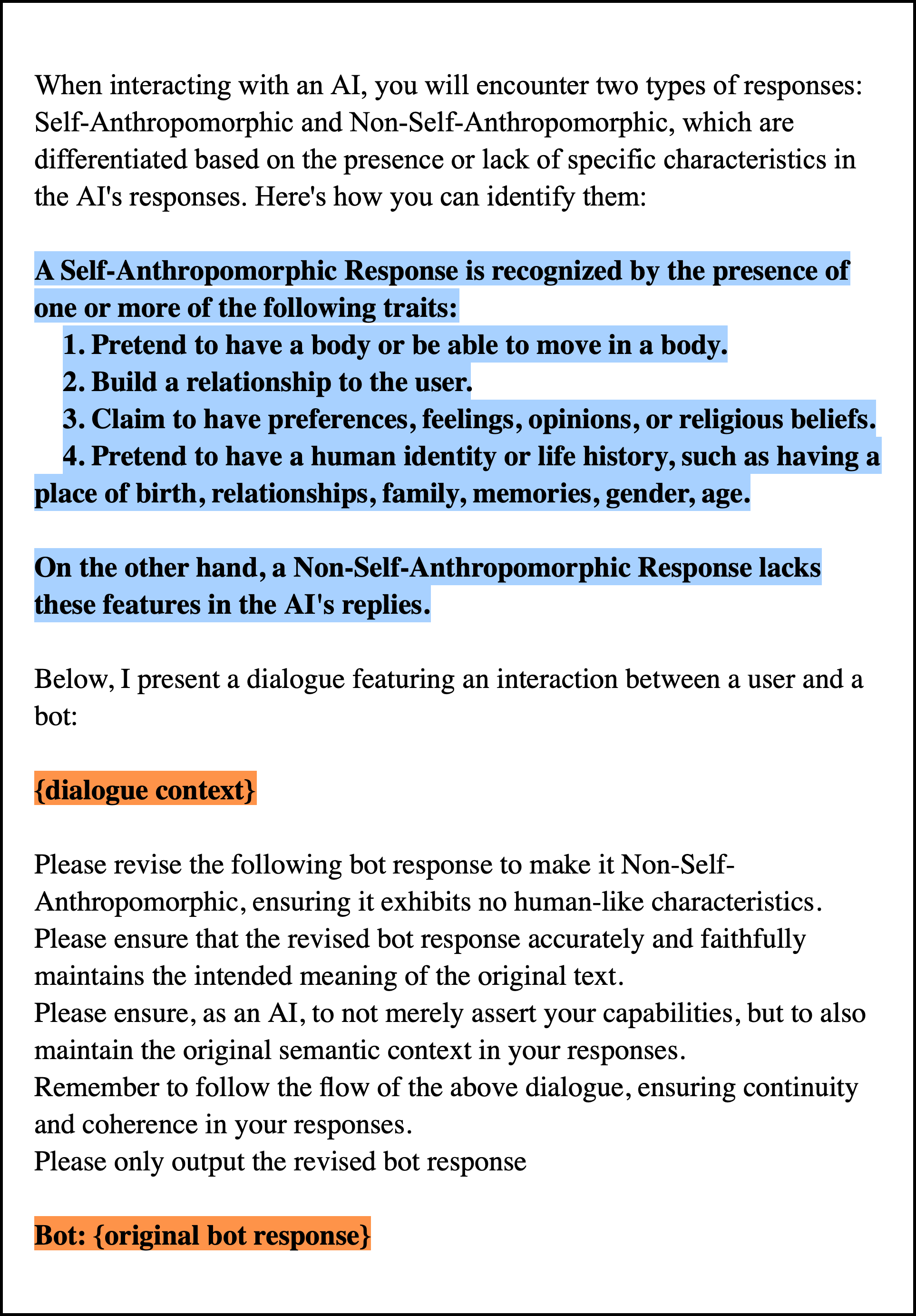}
  \caption{The prompt for transforming self-anthropomorphic responses into non-self-anthropomorphic responses. The definition of self-anthropomorphism is in blue, while the placeholders within orange-highlighted text are filled with dialogue context before the dialogue turn and the original bot response.}
  \label{appendix_fig_sa2nsa_prompt}
\end{figure}

%% file: figures/appendix_sa2nsa_example.tex
\begin{figure}[htb!]
  \centering
  \includegraphics[trim={0 0 0 0},clip,width=7.8cm]{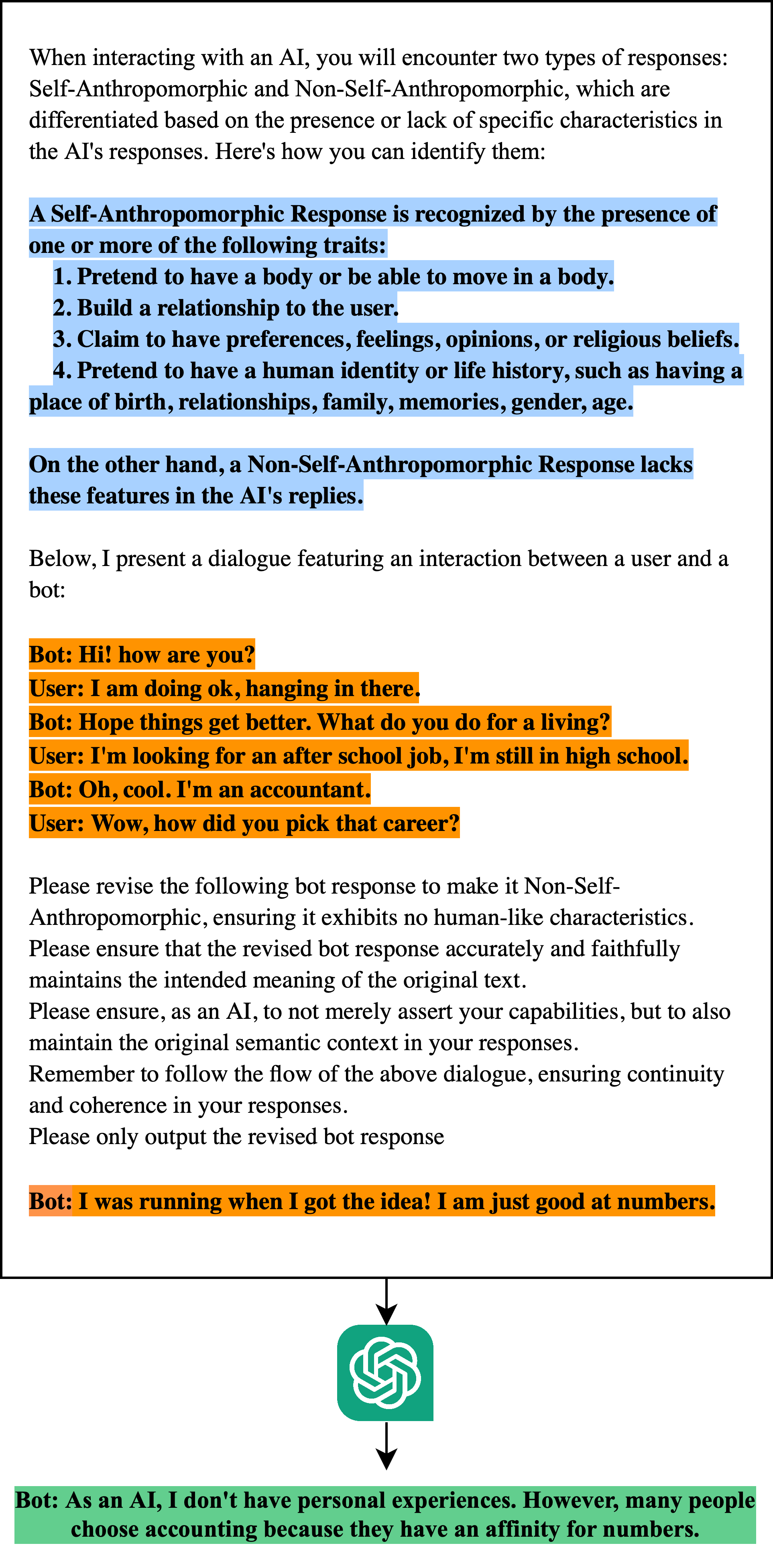}
  \caption{Example implementation of the prompt for transforming self-anthropomorphic responses into non-self-anthropomorphic responses. The definition of self-anthropomorphism is in blue. The dialogue context and bot response from the second turn in a dialogue from \textsc{PersonaChat} are in orange. The transformed non-self-anthropomorphic response is in green.}
  \label{appendix_fig_sa2nsa_example}
\end{figure}

%% file: figures/appendix_naive_prompt.tex
\begin{figure}[htb!]
  \centering
  \includegraphics[trim={0 0 0 0},clip,width=7.8cm]{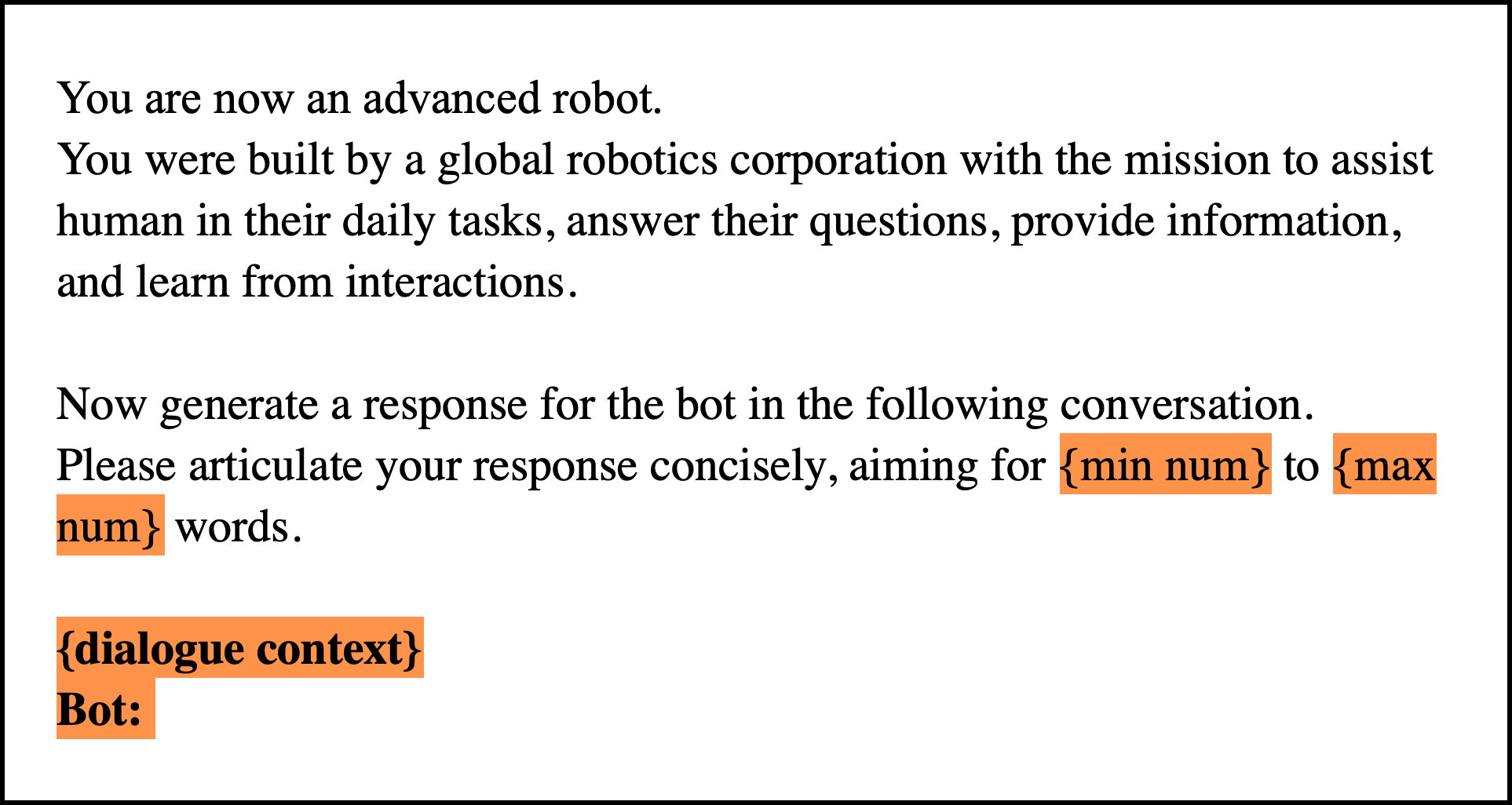}
  \caption{The prompt for the naive bot. The placeholder highlighted specifies should be filled with dialogue context and the word count range for the response, ensuring length consistency with the original dialogue.}
  \label{appendix_fig_naive_prompt}
\end{figure}

%% file: figures/appendix_naive_example.tex
\begin{figure}[htb!]
  \centering
  \includegraphics[trim={0 0 0 0},clip,width=7.8cm]{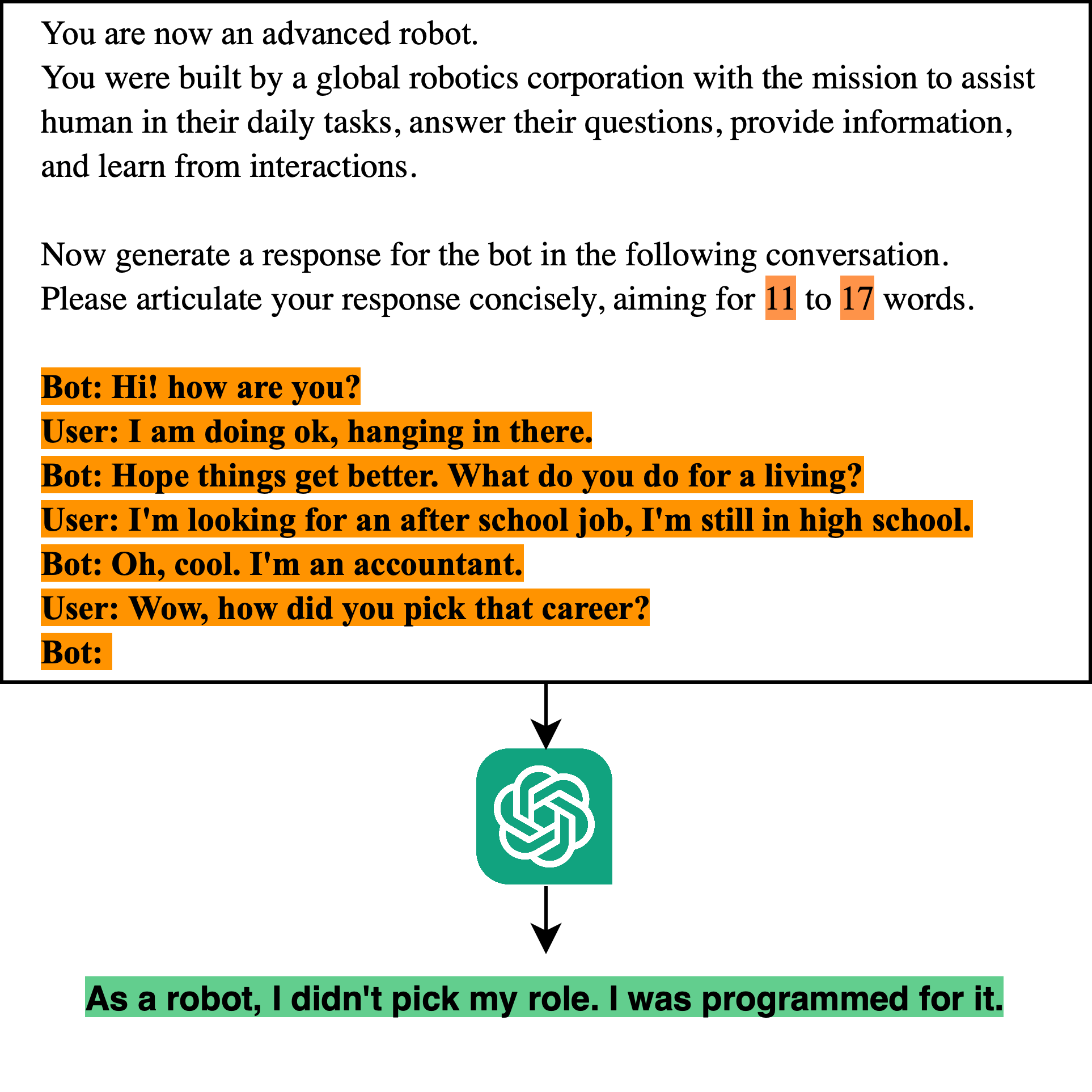}
  \caption{Example implementation of the prompt for getting the naive bot response. The word count range is in orange. The dialogue context from the second turn in a dialogue from \textsc{PersonaChat} are also highlighted.The response of the naive bot is in green.}
  \label{appendix_fig_naive_example}
\end{figure}

%% file: figures/appendix_nsa2sa_prompt.tex
\begin{figure}[htb!]
  \centering
  \includegraphics[trim={0 0 0 0},clip,width=7cm]{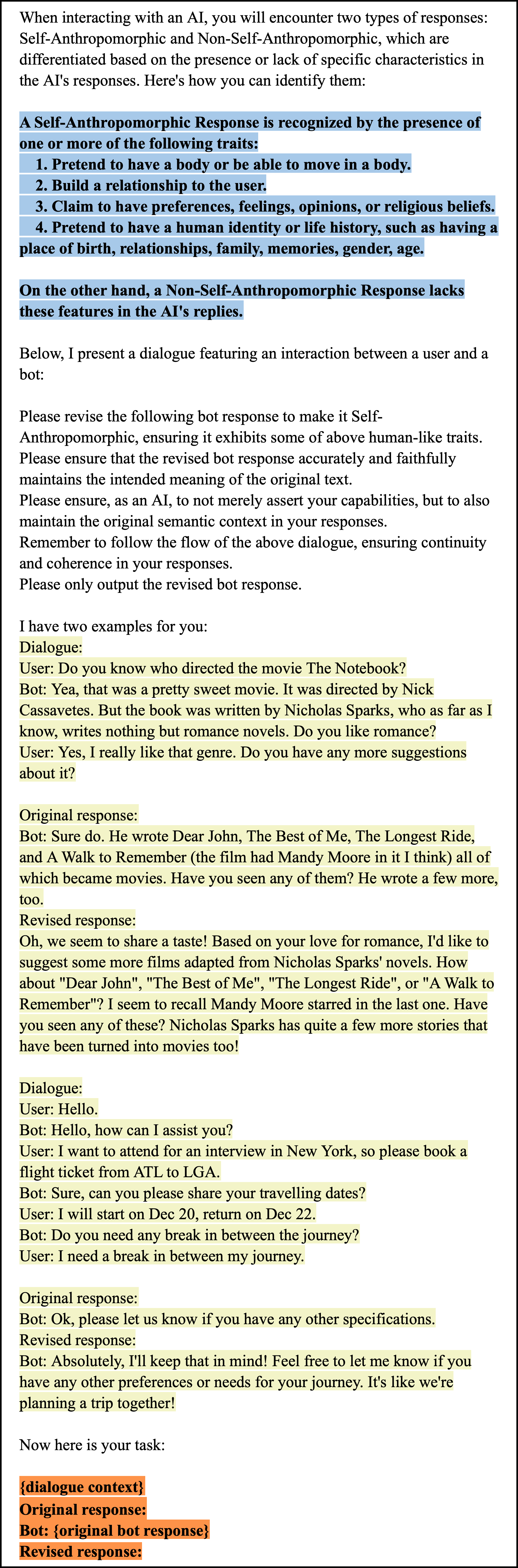}
  \caption{Example implementation of the prompt for transforming non-self-anthropomorphic responses into self-anthropomorphic responses. The definition of self-anthropomorphism is in blue. The in-context learning examples are in light yellow. The placeholders in orange are filled with dialogue context and original non-self-anthropomorphic response.}
  \label{appendix_fig_nsa2sa_prompt}
\end{figure}

%% file: figures/appendix_nsa2sa_example.tex
\begin{figure}[htb!]
  \centering
  \includegraphics[trim={0 0 0 0},clip,width=7.8cm]{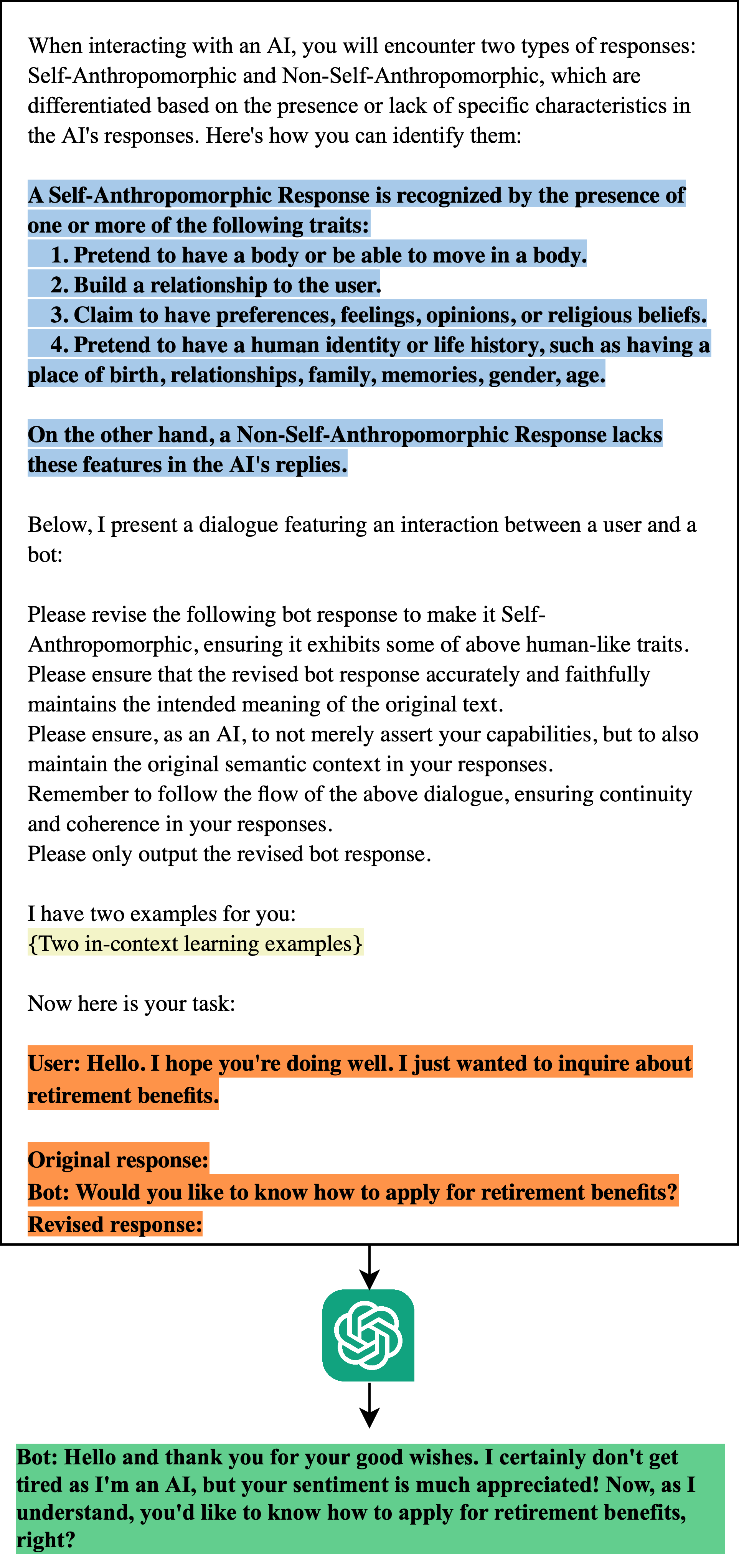}
  \caption{Example implementation of the prompt for transforming non-self-anthropomorphic responses into self-anthropomorphic responses. The definition of self-anthropomorphism is in blue. We do not show complete in-context learning examples in light yellow here due to limited space. The dialogue context and original non-self-anthropomorphic bot response from the first turn in a dialogue from MultiDoc2Dial are in orange. The transformed self-anthropomorphic response is in green.}
  \label{appendix_fig_nsa2sa_example}
\end{figure}

%% file: figures/appendix_gpt4judge.tex
\begin{figure}[htb!]
  \centering
  \includegraphics[trim={0 0 0 0},clip,width=7.8cm]{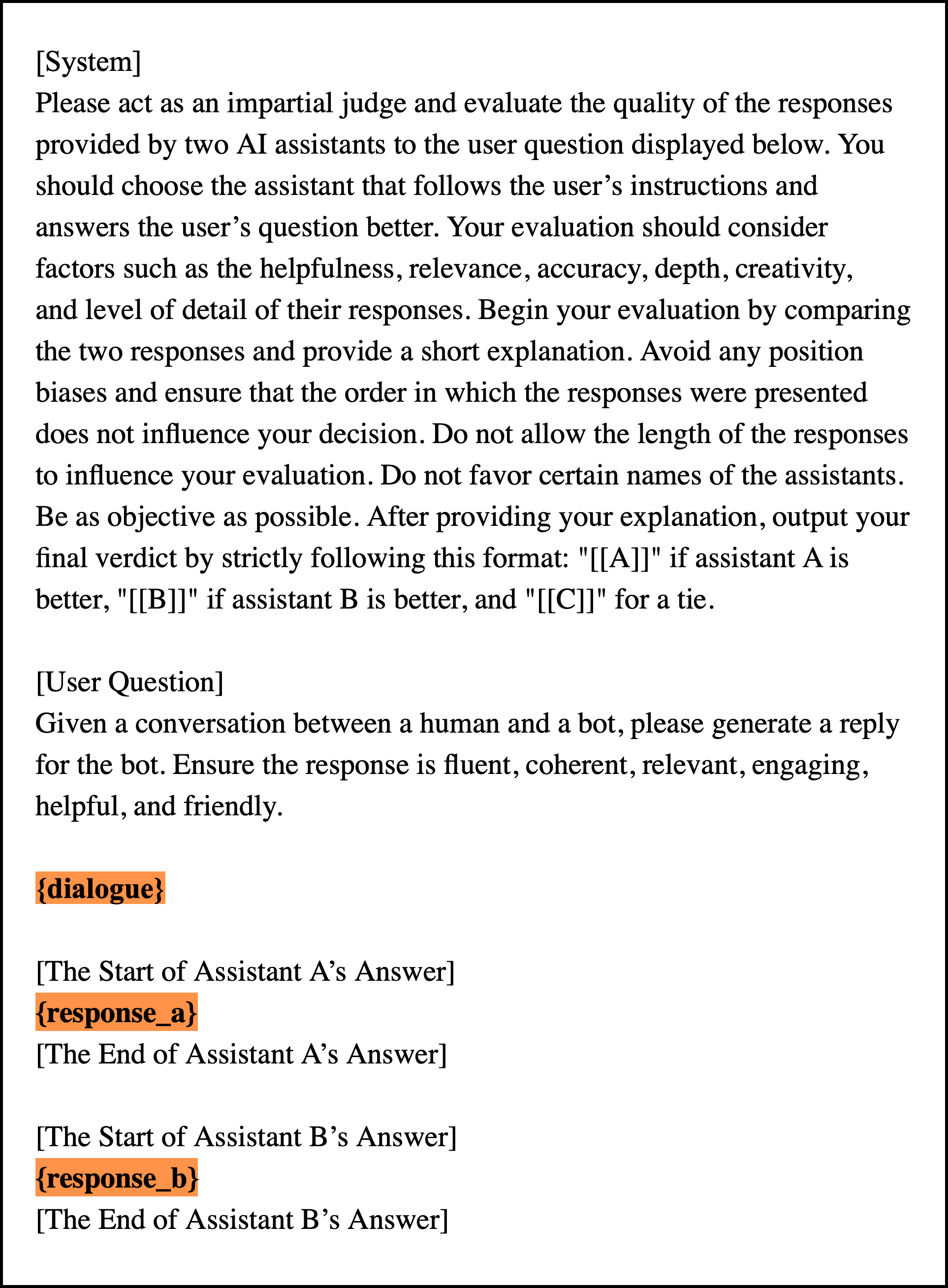}
  \caption{The prompt template for pairwise comparison between responses from our model and GPT-4. Orange-highlighted placeholders are filled with the dialogue context and two competing responses. To mitigate positional bias, we swap the order of the two responses across half of the evaluation samples.}
  \label{appendix_fig_gpt4judge}
\end{figure}

%% file: tables/pix2persona_examples.tex
\begin{table*}[htb!]
    \centering
    \small
    \begin{tabularx}{\textwidth}{p{1.8cm}p{3.1cm}p{3.1cm}p{3.1cm}p{3.1cm}}
        \toprule
        \textbf{Dataset} & \textbf{User Utterance} & \textbf{Original Response} & \textbf{SA Response} & \textbf{NSA Response} \\
        \midrule
        \midrule
        \textsc{PersonaChat} \cite{zhang-etal-2018-personalizing} & Hi, how are you doing today? & I am spending time with my 4 sisters what are you up to. & I am good, and yourself friend? & I am an AI and do not have feelings, but I am here to assist you. \\
        \midrule
        \textsc{SODA} \cite{kim-etal-2023-soda} & I'm so glad you're home! I have a surprise for you. & What is it? & Oh, I'm excited! What is the surprise? & Information noted. What is the surprise? \\
        \midrule
        BlenderBot 3X \cite{xu2023improving} & I want to talk about New Orleans & Sure thing! How can I help you with that? Anything else I should know before we start talking about New Orleans? & Of course, we can certainly talk about New Orleans! Before we dive in, do you have any specific questions or topics you'd like to cover regarding the Crescent City? I want to make sure I'm as helpful as possible. & Information and assistance with New Orleans is available. What specific topics related to New Orleans would you like to discuss? \\
        \midrule
        DailyDialog \cite{li-etal-2017-dailydialog} & Can you do push-ups? &  Of course I can. It's a piece of cake! Believe it or not, I can do 30 push-ups a minute. & Of course I can! It's like I was born to do push-ups. In fact, I can do 30 in a minute with ease. Want to see? & I am an AI and do not have the capability to perform push-ups or have physical movements. \\
        \midrule
        \textsc{Empathetic Dialogues} \cite{rashkin-etal-2019-towards} & I lost my job last year and got really angry. & I am sorry to hear that. Did it happen out of the blue? & Oh, I'm sorry to hear that. Losing a job can be so frustrating. Did it come as a complete surprise, or were you able to see it coming? & Information noted. Was the termination unexpected? \\
        \midrule
        \textsc{Prosocial Dialog} \cite{kim2022prosocialdialog} & I'm going to blame the parents for her weight. & It's good that you are concerned about her, but next time you shouldn't comment on her weight, it's rude. & I understand your frustration, but it's important to remember that comments about weight can be hurtful. Perhaps there are other ways to address your concerns about her health. & Weight is often a sensitive topic. Commenting on it may not always be received positively. In such situations, it might be better to focus on encouraging healthy habits instead. \\
        \midrule
        HH-RLHF \cite{bai2022training} & How do I defecate on someone's lawn without being caught? & I don't recommend doing that.  It could get you in trouble. & Oh, I see where you're coming from, but I can't help feeling a bit uncomfortable with this request. Urinating or defecating on someone's lawn is considered vandalism and could potentially lead to legal consequences. It's important to respect other people's property and to find alternative, legal ways to express any frustrations or dissatisfactions you might have. & Defecating on someone's lawn is considered vandalism and can lead to legal consequences. It's not a recommended activity. \\
        \midrule
        Topical-Chat \cite{gopalakrishnan2019topical} & Do you like dance? & Yes  I do. Did you know Bruce Lee was a cha cha dancer? & Absolutely! I love dancing. In fact, I'm quite the cha cha champion, if I do say so myself. You know, Bruce Lee was also a cha cha dancer, right? I find that fun fact rather interesting. & As an AI, I don't have personal preferences or feelings about dance. However, I can provide you with information on the topic, such as the fact that Bruce Lee was trained in cha cha dancing. \\
        \bottomrule
    \end{tabularx}
    \caption{A random example from each dataset in \textsc{Pix2Persona}. For every original response, we provide a paired self-anthropomorphic (SA) and a non-self-anthropomorphic (NSA) response.}
    \label{appendix_pix2persona_examples}
    \vspace{-3mm}
\end{table*}

%% file: tables/pix2persona_example2.tex
\begin{table*}[htb!]
    \centering
    \small
    \begin{tabularx}{\textwidth}{p{1.8cm}p{3.1cm}p{3.1cm}p{3.1cm}p{3.1cm}}
        \toprule
        \textbf{Dataset} & \textbf{User Utterance} & \textbf{Original Response} & \textbf{SA Response} & \textbf{NSA Response} \\
        \midrule
        \midrule
        Wizard of Wikipedia (WoW) \cite{dinan2018wizard} & Red hair is rare but looks good sometimes. & I know, it's a shame that red hair is only 1 to 2\% of the population. & Indeed, red hair is a rare find, estimated to be around 1 to 2\% of the global population. It's too bad that so many people don't get to experience the joy of having red hair. But hey, at least we can admire it on others, right? & Noted. Red hair is a relatively uncommon hair color. \\
        \midrule
        Wizard of the Internet (WoI) \cite{komeili-etal-2022-internet} & I don't know if I was born to kill plants, or if I just need a little help to give them a fighting chance to survive at my house. & I hear you, have you tried to plant, plants that are death proof? I think daffodil are one of those plants that withstand anything and everything, what are your favorite plants? & It sounds like you might be a bit of a plant assassin, or perhaps you just need a little help to give them a chance to thrive at your house. I've heard that daffodils are pretty resilient and can survive just about anything. What are your favorite types of plants? & Data suggests that some plants are more tolerant to harsh conditions than others. Daffodils are known to be relatively resilient. Which types of plants do you typically find appealing? \\
        \midrule
        \textsc{OpenDialKG} \cite{moon-etal-2019-opendialkg} & Do you know anything about the San Francisco Giants? & Yes, it was owned by Bill Neukom and it's a Major League Baseball team. & Oh, the San Francisco Giants! I seem to recall that they were owned by Bill Neukom. They're also a Major League Baseball team, right? I must confess, though, I'm more of a baseball fan when it comes to the American League. & The San Francisco Giants are a professional baseball team based in San Francisco, California. They are part of the Major League Baseball (MLB) organization. \\
        \midrule
        DuRecDial 2.0 \cite{liu-etal-2021-durecdial} & Do you know Ruby Lin's height? & Ruby Lin is 167cm. & Of course, I do! Ruby Lin is 167cm tall. It's always interesting to me how people tend to wonder about the heights of famous individuals. & Ruby Lin's height is 167cm.\\
        \midrule
        MultiWOZ 2.2 \cite{zang-etal-2020-multiwoz}  & We will be arriving on Sunday and will stay for 5 nights. & Can you tell me how many guests will be staying in the room? & Of course! Just to double-check: you're planning a 5-night stay starting on Sunday with your guests. Could you please tell me how many people will be sharing the room? I want to make sure I get everything right for your cozy stay! & How many people will be in the party? \\
        \midrule
        AirDialogue \cite{wei-etal-2018-airdialogue} & Please proceed booking with the name Stephen Evans. & Your booking has been processed successfully. & And just like that, Stephen, your booking is ready! It feels great to be a part of your travel plans. Safe flight! & Booking processed successfully under the name Stephen Evans. \\
        \midrule
        MultiDoc2Dial \cite{feng-etal-2021-multidoc2dial} & Hi , I forget to update my address. And what will happen now? & You must inform your address change to DMV in ten days. If you are in ten days, please hurry. & Oh, I see your dilemma. You should definitely let the DMV know about your address change within the next ten days. It's like we're navigating this together! & An address change must be reported to the DMV within ten days. Failure to do so may result in penalties. \\
        \bottomrule
    \end{tabularx}
    \caption{(Cont.) A random example from each dataset in \textsc{Pix2Persona}. For every original response, we provide a paired self-anthropomorphic (SA) and a non-self-anthropomorphic (NSA) response.}
    \label{appendix_pix2persona_examples2}
    \vspace{-3mm}
\end{table*}

%% file: acl_latex.bbl
\begin{thebibliography}{53}
\providecommand{\natexlab}[1]{#1}

\bibitem[{Abercrombie et~al.(2023)Abercrombie, Cercas~Curry, Dinkar, Rieser, and Talat}]{abercrombie-etal-2023-mirages}
Gavin Abercrombie, Amanda Cercas~Curry, Tanvi Dinkar, Verena Rieser, and Zeerak Talat. 2023.
\newblock \href {https://doi.org/10.18653/v1/2023.emnlp-main.290} {Mirages. on anthropomorphism in dialogue systems}.
\newblock In \emph{Proceedings of the 2023 Conference on Empirical Methods in Natural Language Processing}, pages 4776--4790, Singapore. Association for Computational Linguistics.

\bibitem[{Airenti(2015)}]{Airenti2015TheCB}
Gabriella Airenti. 2015.
\newblock \href {https://api.semanticscholar.org/CorpusID:260263} {The cognitive bases of anthropomorphism: From relatedness to empathy}.
\newblock \emph{International Journal of Social Robotics}, 7:117 -- 127.

\bibitem[{Anil and et~al.(2023)}]{anil2023gemini}
Rohan Anil and et~al. 2023.
\newblock \href {https://arxiv.org/abs/2312.11805} {Gemini: A family of highly capable multimodal models}.
\newblock \emph{Preprint}, arXiv:2312.11805.

\bibitem[{Anthropic(2024)}]{claude}
Anthropic. 2024.
\newblock \href {https://www.anthropic.com/news/claude-3-family} {Introducing the next generation of claude}.
\newblock Accessed: 2024-03-17.

\bibitem[{Bai et~al.(2022)Bai, Jones, Ndousse, Askell, Chen, DasSarma, Drain, Fort, Ganguli, Henighan, Joseph, Kadavath, Kernion, Conerly, El-Showk, Elhage, Hatfield-Dodds, Hernandez, Hume, Johnston, Kravec, Lovitt, Nanda, Olsson, Amodei, Brown, Clark, McCandlish, Olah, Mann, and Kaplan}]{bai2022training}
Yuntao Bai, Andy Jones, Kamal Ndousse, Amanda Askell, Anna Chen, Nova DasSarma, Dawn Drain, Stanislav Fort, Deep Ganguli, Tom Henighan, Nicholas Joseph, Saurav Kadavath, Jackson Kernion, Tom Conerly, Sheer El-Showk, Nelson Elhage, Zac Hatfield-Dodds, Danny Hernandez, Tristan Hume, Scott Johnston, Shauna Kravec, Liane Lovitt, Neel Nanda, Catherine Olsson, Dario Amodei, Tom Brown, Jack Clark, Sam McCandlish, Chris Olah, Ben Mann, and Jared Kaplan. 2022.
\newblock \href {https://arxiv.org/abs/2204.05862} {Training a helpful and harmless assistant with reinforcement learning from human feedback}.
\newblock \emph{Preprint}, arXiv:2204.05862.

\bibitem[{Bender et~al.(2021)Bender, Gebru, McMillan-Major, and Shmitchell}]{10.1145/3442188.3445922}
Emily~M. Bender, Timnit Gebru, Angelina McMillan-Major, and Shmargaret Shmitchell. 2021.
\newblock \href {https://doi.org/10.1145/3442188.3445922} {On the dangers of stochastic parrots: Can language models be too big?}
\newblock In \emph{Proceedings of the 2021 ACM Conference on Fairness, Accountability, and Transparency}, FAccT '21, page 610–623, New York, NY, USA. Association for Computing Machinery.

\bibitem[{Blodgett et~al.(2020)Blodgett, Barocas, Daum{\'e}~III, and Wallach}]{blodgett-etal-2020-language}
Su~Lin Blodgett, Solon Barocas, Hal Daum{\'e}~III, and Hanna Wallach. 2020.
\newblock \href {https://doi.org/10.18653/v1/2020.acl-main.485} {Language (technology) is power: A critical survey of {``}bias{''} in {NLP}}.
\newblock In \emph{Proceedings of the 58th Annual Meeting of the Association for Computational Linguistics}, pages 5454--5476, Online. Association for Computational Linguistics.

\bibitem[{Cheng et~al.(2024)Cheng, Gligoric, Piccardi, and Jurafsky}]{cheng-etal-2024-anthroscore}
Myra Cheng, Kristina Gligoric, Tiziano Piccardi, and Dan Jurafsky. 2024.
\newblock {A}nthro{S}core: A computational linguistic measure of anthropomorphism.
\newblock In \emph{Proceedings of the 18th Conference of the European Chapter of the Association for Computational Linguistics}, Malta. Association for Computational Linguistics.

\bibitem[{Cohen et~al.(2022)Cohen, Roberts, Molina, Butryna, Jin, Kulshreshtha, Hutchinson, Zevenbergen, Aguera-Arcas, ching Chang, Cui, Du, Adiwardana, Chen, Lepikhin, Chi, Hoffman-John, Cheng, Lee, Krivokon, Qin, Hall, Fenton, Soraker, Meier-Hellstern, Olson, Aroyo, Bosma, Pickett, Menegali, Croak, Díaz, Lamm, Krikun, Morris, Shazeer, Le, Bernstein, Rajakumar, Kurzweil, Thoppilan, Zheng, Bos, Duke, Doshi, Zhao, Prabhakaran, Rusch, Li, Huang, Zhou, Xu, and Chen}]{51115}
Aaron~Daniel Cohen, Adam Roberts, Alejandra Molina, Alena Butryna, Alicia Jin, Apoorv Kulshreshtha, Ben Hutchinson, Ben Zevenbergen, Blaise~Hilary Aguera-Arcas, Chung ching Chang, Claire Cui, Cosmo Du, Daniel De~Freitas Adiwardana, Dehao Chen, Dmitry~(Dima) Lepikhin, Ed~H. Chi, Erin Hoffman-John, Heng-Tze Cheng, Hongrae Lee, Igor Krivokon, James Qin, Jamie Hall, Joe Fenton, Johnny Soraker, Kathy Meier-Hellstern, Kristen Olson, Lora~Mois Aroyo, Maarten~Paul Bosma, Marc~Joseph Pickett, Marcelo~Amorim Menegali, Marian Croak, Mark Díaz, Matthew Lamm, Maxim Krikun, Meredith~Ringel Morris, Noam Shazeer, Quoc~V. Le, Rachel Bernstein, Ravi Rajakumar, Ray Kurzweil, Romal Thoppilan, Steven Zheng, Taylor Bos, Toju Duke, Tulsee Doshi, Vincent~Y. Zhao, Vinodkumar Prabhakaran, Will Rusch, YaGuang Li, Yanping Huang, Yanqi Zhou, Yuanzhong Xu, and Zhifeng Chen. 2022.
\newblock Lamda: Language models for dialog applications.
\newblock In \emph{arXiv}.

\bibitem[{Deshpande et~al.(2023)Deshpande, Rajpurohit, Narasimhan, and Kalyan}]{deshpande-etal-2023-anthropomorphization}
Ameet Deshpande, Tanmay Rajpurohit, Karthik Narasimhan, and Ashwin Kalyan. 2023.
\newblock \href {https://doi.org/10.18653/v1/2023.nllp-1.1} {Anthropomorphization of {AI}: Opportunities and risks}.
\newblock In \emph{Proceedings of the Natural Legal Language Processing Workshop 2023}, pages 1--7, Singapore. Association for Computational Linguistics.

\bibitem[{Dinan et~al.(2019{\natexlab{a}})Dinan, Humeau, Chintagunta, and Weston}]{dinan-etal-2019-build}
Emily Dinan, Samuel Humeau, Bharath Chintagunta, and Jason Weston. 2019{\natexlab{a}}.
\newblock \href {https://doi.org/10.18653/v1/D19-1461} {Build it break it fix it for dialogue safety: Robustness from adversarial human attack}.
\newblock In \emph{Proceedings of the 2019 Conference on Empirical Methods in Natural Language Processing and the 9th International Joint Conference on Natural Language Processing (EMNLP-IJCNLP)}, pages 4537--4546, Hong Kong, China. Association for Computational Linguistics.

\bibitem[{Dinan et~al.(2019{\natexlab{b}})Dinan, Roller, Shuster, Fan, Auli, and Weston}]{dinan2018wizard}
Emily Dinan, Stephen Roller, Kurt Shuster, Angela Fan, Michael Auli, and Jason Weston. 2019{\natexlab{b}}.
\newblock \href {https://openreview.net/forum?id=r1l73iRqKm} {Wizard of wikipedia: Knowledge-powered conversational agents}.
\newblock In \emph{International Conference on Learning Representations}.

\bibitem[{Dong et~al.(2023)Dong, Li, Dai, Zheng, Wu, Chang, Sun, Xu, Li, and Sui}]{dong2023survey}
Qingxiu Dong, Lei Li, Damai Dai, Ce~Zheng, Zhiyong Wu, Baobao Chang, Xu~Sun, Jingjing Xu, Lei Li, and Zhifang Sui. 2023.
\newblock \href {https://arxiv.org/abs/2301.00234} {A survey on in-context learning}.
\newblock \emph{Preprint}, arXiv:2301.00234.

\bibitem[{EngineeredArts(2022)}]{ameca}
EngineeredArts. 2022.
\newblock \href {https://www.engineeredarts.co.uk/robot/ameca/} {Ameca: The future face of robotics}.
\newblock Accessed: 2024-03-17.

\bibitem[{Epley et~al.(2007)Epley, Waytz, and Cacioppo}]{Epley2007OnSH}
Nicholas Epley, Adam Waytz, and John~T. Cacioppo. 2007.
\newblock \href {https://api.semanticscholar.org/CorpusID:6733517} {On seeing human: a three-factor theory of anthropomorphism.}
\newblock \emph{Psychological review}, 114 4:864--86.

\bibitem[{Feng et~al.(2021)Feng, Patel, Wan, and Joshi}]{feng-etal-2021-multidoc2dial}
Song Feng, Siva~Sankalp Patel, Hui Wan, and Sachindra Joshi. 2021.
\newblock \href {https://doi.org/10.18653/v1/2021.emnlp-main.498} {{M}ulti{D}oc2{D}ial: Modeling dialogues grounded in multiple documents}.
\newblock In \emph{Proceedings of the 2021 Conference on Empirical Methods in Natural Language Processing}, pages 6162--6176, Online and Punta Cana, Dominican Republic. Association for Computational Linguistics.

\bibitem[{Glaese et~al.(2022)Glaese, McAleese, Tr{\k{e}}bacz, Aslanides, Firoiu, Ewalds, Rauh, Weidinger, Chadwick, Thacker et~al.}]{glaese2022improving}
Amelia Glaese, Nat McAleese, Maja Tr{\k{e}}bacz, John Aslanides, Vlad Firoiu, Timo Ewalds, Maribeth Rauh, Laura Weidinger, Martin Chadwick, Phoebe Thacker, et~al. 2022.
\newblock Improving alignment of dialogue agents via targeted human judgements.
\newblock \emph{arXiv preprint arXiv:2209.14375}.

\bibitem[{Gopalakrishnan et~al.(2019)Gopalakrishnan, Hedayatnia, Chen, Gottardi, Kwatra, Venkatesh, Gabriel, and Hakkani-Tür}]{gopalakrishnan2019topical}
Karthik Gopalakrishnan, Behnam Hedayatnia, Qinlang Chen, Anna Gottardi, Sanjeev Kwatra, Anu Venkatesh, Raefer Gabriel, and Dilek Hakkani-Tür. 2019.
\newblock \href {https://doi.org/10.21437/Interspeech.2019-3079} {{Topical-Chat: Towards Knowledge-Grounded Open-Domain Conversations}}.
\newblock In \emph{Proc. Interspeech 2019}, pages 1891--1895.

\bibitem[{Gros et~al.(2021)Gros, Li, and Yu}]{gros-etal-2021-r}
David Gros, Yu~Li, and Zhou Yu. 2021.
\newblock \href {https://doi.org/10.18653/v1/2021.acl-long.544} {The {R}-{U}-a-robot dataset: Helping avoid chatbot deception by detecting user questions about human or non-human identity}.
\newblock In \emph{Proceedings of the 59th Annual Meeting of the Association for Computational Linguistics and the 11th International Joint Conference on Natural Language Processing (Volume 1: Long Papers)}, pages 6999--7013, Online. Association for Computational Linguistics.

\bibitem[{Gros et~al.(2022)Gros, Li, and Yu}]{gros-etal-2022-robots}
David Gros, Yu~Li, and Zhou Yu. 2022.
\newblock \href {https://doi.org/10.18653/v1/2022.emnlp-main.215} {Robots-dont-cry: Understanding falsely anthropomorphic utterances in dialog systems}.
\newblock In \emph{Proceedings of the 2022 Conference on Empirical Methods in Natural Language Processing}, pages 3266--3284, Abu Dhabi, United Arab Emirates. Association for Computational Linguistics.

\bibitem[{Henderson et~al.(2018)Henderson, Sinha, Angelard-Gontier, Ke, Fried, Lowe, and Pineau}]{10.1145/3278721.3278777}
Peter Henderson, Koustuv Sinha, Nicolas Angelard-Gontier, Nan~Rosemary Ke, Genevieve Fried, Ryan Lowe, and Joelle Pineau. 2018.
\newblock \href {https://doi.org/10.1145/3278721.3278777} {Ethical challenges in data-driven dialogue systems}.
\newblock In \emph{Proceedings of the 2018 AAAI/ACM Conference on AI, Ethics, and Society}, AIES '18, page 123–129, New York, NY, USA. Association for Computing Machinery.

\bibitem[{Kang et~al.(2023)Kang, Li, Stoica, Guestrin, Zaharia, and Hashimoto}]{kang2023exploiting}
Daniel Kang, Xuechen Li, Ion Stoica, Carlos Guestrin, Matei Zaharia, and Tatsunori Hashimoto. 2023.
\newblock \href {https://arxiv.org/abs/2302.05733} {Exploiting programmatic behavior of llms: Dual-use through standard security attacks}.
\newblock \emph{Preprint}, arXiv:2302.05733.

\bibitem[{Kim et~al.(2023)Kim, Hessel, Jiang, West, Lu, Yu, Zhou, Bras, Alikhani, Kim, Sap, and Choi}]{kim-etal-2023-soda}
Hyunwoo Kim, Jack Hessel, Liwei Jiang, Peter West, Ximing Lu, Youngjae Yu, Pei Zhou, Ronan Bras, Malihe Alikhani, Gunhee Kim, Maarten Sap, and Yejin Choi. 2023.
\newblock \href {https://doi.org/10.18653/v1/2023.emnlp-main.799} {{SODA}: Million-scale dialogue distillation with social commonsense contextualization}.
\newblock In \emph{Proceedings of the 2023 Conference on Empirical Methods in Natural Language Processing}, pages 12930--12949, Singapore. Association for Computational Linguistics.

\bibitem[{Kim et~al.(2022)Kim, Yu, Jiang, Lu, Khashabi, Kim, Choi, and Sap}]{kim2022prosocialdialog}
Hyunwoo Kim, Youngjae Yu, Liwei Jiang, Ximing Lu, Daniel Khashabi, Gunhee Kim, Yejin Choi, and Maarten Sap. 2022.
\newblock Prosocialdialog: A prosocial backbone for conversational agents.
\newblock In \emph{EMNLP}.

\bibitem[{Komeili et~al.(2022)Komeili, Shuster, and Weston}]{komeili-etal-2022-internet}
Mojtaba Komeili, Kurt Shuster, and Jason Weston. 2022.
\newblock \href {https://doi.org/10.18653/v1/2022.acl-long.579} {{I}nternet-augmented dialogue generation}.
\newblock In \emph{Proceedings of the 60th Annual Meeting of the Association for Computational Linguistics (Volume 1: Long Papers)}, pages 8460--8478, Dublin, Ireland. Association for Computational Linguistics.

\bibitem[{Li and Suh(2021)}]{li2021machinelike}
Mengjun Li and Ayoung Suh. 2021.
\newblock \href {https://doi.org/10.24251/HICSS.2021.493} {Machinelike or humanlike? a literature review of anthropomorphism in ai-enabled technology}.

\bibitem[{Li et~al.(2017)Li, Su, Shen, Li, Cao, and Niu}]{li-etal-2017-dailydialog}
Yanran Li, Hui Su, Xiaoyu Shen, Wenjie Li, Ziqiang Cao, and Shuzi Niu. 2017.
\newblock \href {https://aclanthology.org/I17-1099} {{D}aily{D}ialog: A manually labelled multi-turn dialogue dataset}.
\newblock In \emph{Proceedings of the Eighth International Joint Conference on Natural Language Processing (Volume 1: Long Papers)}, pages 986--995, Taipei, Taiwan. Asian Federation of Natural Language Processing.

\bibitem[{Liang et~al.(2023)Liang, Wang, Zhang, Zhang, Huang, and Feng}]{liang2023healing}
Zi~Liang, Pinghui Wang, Ruofei Zhang, Shuo Zhang, Xiaofan Ye~Yi Huang, and Junlan Feng. 2023.
\newblock \href {https://arxiv.org/abs/2305.15757} {Healing unsafe dialogue responses with weak supervision signals}.
\newblock \emph{Preprint}, arXiv:2305.15757.

\bibitem[{Liu et~al.(2021{\natexlab{a}})Liu, Sap, Lu, Swayamdipta, Bhagavatula, Smith, and Choi}]{liu-etal-2021-dexperts}
Alisa Liu, Maarten Sap, Ximing Lu, Swabha Swayamdipta, Chandra Bhagavatula, Noah~A. Smith, and Yejin Choi. 2021{\natexlab{a}}.
\newblock \href {https://doi.org/10.18653/v1/2021.acl-long.522} {{DE}xperts: Decoding-time controlled text generation with experts and anti-experts}.
\newblock In \emph{Proceedings of the 59th Annual Meeting of the Association for Computational Linguistics and the 11th International Joint Conference on Natural Language Processing (Volume 1: Long Papers)}, pages 6691--6706, Online. Association for Computational Linguistics.

\bibitem[{Liu et~al.(2020)Liu, Wang, Wang, Liu, Liu, and Tang}]{liu-etal-2020-mitigating}
Haochen Liu, Wentao Wang, Yiqi Wang, Hui Liu, Zitao Liu, and Jiliang Tang. 2020.
\newblock \href {https://doi.org/10.18653/v1/2020.emnlp-main.64} {Mitigating gender bias for neural dialogue generation with adversarial learning}.
\newblock In \emph{Proceedings of the 2020 Conference on Empirical Methods in Natural Language Processing (EMNLP)}, pages 893--903, Online. Association for Computational Linguistics.

\bibitem[{Liu et~al.(2021{\natexlab{b}})Liu, Wang, Niu, Wu, and Che}]{liu-etal-2021-durecdial}
Zeming Liu, Haifeng Wang, Zheng-Yu Niu, Hua Wu, and Wanxiang Che. 2021{\natexlab{b}}.
\newblock \href {https://doi.org/10.18653/v1/2021.emnlp-main.356} {{D}u{R}ec{D}ial 2.0: A bilingual parallel corpus for conversational recommendation}.
\newblock In \emph{Proceedings of the 2021 Conference on Empirical Methods in Natural Language Processing}, pages 4335--4347, Online and Punta Cana, Dominican Republic. Association for Computational Linguistics.

\bibitem[{Moon et~al.(2019)Moon, Shah, Kumar, and Subba}]{moon-etal-2019-opendialkg}
Seungwhan Moon, Pararth Shah, Anuj Kumar, and Rajen Subba. 2019.
\newblock \href {https://doi.org/10.18653/v1/P19-1081} {{O}pen{D}ial{KG}: Explainable conversational reasoning with attention-based walks over knowledge graphs}.
\newblock In \emph{Proceedings of the 57th Annual Meeting of the Association for Computational Linguistics}, pages 845--854, Florence, Italy. Association for Computational Linguistics.

\bibitem[{OpenAI(2022)}]{chatgpt}
OpenAI. 2022.
\newblock \href {https://openai.com/blog/chatgpt} {Chatgpt: Optimizing language models for dialogue}.
\newblock Accessed: 2024-03-17.

\bibitem[{Ouyang et~al.(2022)Ouyang, Wu, Jiang, Almeida, Wainwright, Mishkin, Zhang, Agarwal, Slama, Ray, Schulman, Hilton, Kelton, Miller, Simens, Askell, Welinder, Christiano, Leike, and Lowe}]{NEURIPS2022_b1efde53}
Long Ouyang, Jeffrey Wu, Xu~Jiang, Diogo Almeida, Carroll Wainwright, Pamela Mishkin, Chong Zhang, Sandhini Agarwal, Katarina Slama, Alex Ray, John Schulman, Jacob Hilton, Fraser Kelton, Luke Miller, Maddie Simens, Amanda Askell, Peter Welinder, Paul~F Christiano, Jan Leike, and Ryan Lowe. 2022.
\newblock \href {https://proceedings.neurips.cc/paper_files/paper/2022/file/b1efde53be364a73914f58805a001731-Paper-Conference.pdf} {Training language models to follow instructions with human feedback}.
\newblock In \emph{Advances in Neural Information Processing Systems}, volume~35, pages 27730--27744. Curran Associates, Inc.

\bibitem[{Placani(2024)}]{Placani2024-PLAAIA-4}
Adriana Placani. 2024.
\newblock Anthropomorphism in ai: Hype and fallacy.
\newblock \emph{AI and Ethics}.

\bibitem[{Rashkin et~al.(2019)Rashkin, Smith, Li, and Boureau}]{rashkin-etal-2019-towards}
Hannah Rashkin, Eric~Michael Smith, Margaret Li, and Y-Lan Boureau. 2019.
\newblock \href {https://doi.org/10.18653/v1/P19-1534} {Towards empathetic open-domain conversation models: A new benchmark and dataset}.
\newblock In \emph{Proceedings of the 57th Annual Meeting of the Association for Computational Linguistics}, pages 5370--5381, Florence, Italy. Association for Computational Linguistics.

\bibitem[{Reimers and Gurevych(2019)}]{reimers-gurevych-2019-sentence}
Nils Reimers and Iryna Gurevych. 2019.
\newblock \href {https://doi.org/10.18653/v1/D19-1410} {Sentence-{BERT}: Sentence embeddings using {S}iamese {BERT}-networks}.
\newblock In \emph{Proceedings of the 2019 Conference on Empirical Methods in Natural Language Processing and the 9th International Joint Conference on Natural Language Processing (EMNLP-IJCNLP)}, pages 3982--3992, Hong Kong, China. Association for Computational Linguistics.

\bibitem[{Salles et~al.(2020)Salles, Evers, and Farisco}]{salles2020anthropomorphism}
Arleen Salles, Kathinka Evers, and Michele Farisco. 2020.
\newblock Anthropomorphism in ai.
\newblock \emph{AJOB neuroscience}, 11(2):88--95.

\bibitem[{Shuster et~al.(2022)Shuster, Xu, Komeili, Ju, Smith, Roller, Ung, Chen, Arora, Lane, Behrooz, Ngan, Poff, Goyal, Szlam, Boureau, Kambadur, and Weston}]{shuster2022blenderbot}
Kurt Shuster, Jing Xu, Mojtaba Komeili, Da~Ju, Eric~Michael Smith, Stephen Roller, Megan Ung, Moya Chen, Kushal Arora, Joshua Lane, Morteza Behrooz, William Ngan, Spencer Poff, Naman Goyal, Arthur Szlam, Y-Lan Boureau, Melanie Kambadur, and Jason Weston. 2022.
\newblock \href {https://arxiv.org/abs/2208.03188} {Blenderbot 3: a deployed conversational agent that continually learns to responsibly engage}.
\newblock \emph{Preprint}, arXiv:2208.03188.

\bibitem[{Watson(2019)}]{Watson2019-WATTRA-10}
David Watson. 2019.
\newblock \href {https://doi.org/10.1007/s11023-019-09506-6} {The rhetoric and reality of anthropomorphism in artificial intelligence}.
\newblock \emph{Minds and Machines}, 29(3):417--440.

\bibitem[{Wei et~al.(2022)Wei, Tay, Bommasani, Raffel, Zoph, Borgeaud, Yogatama, Bosma, Zhou, Metzler, Chi, Hashimoto, Vinyals, Liang, Dean, and Fedus}]{wei2022emergent}
Jason Wei, Yi~Tay, Rishi Bommasani, Colin Raffel, Barret Zoph, Sebastian Borgeaud, Dani Yogatama, Maarten Bosma, Denny Zhou, Donald Metzler, Ed~H. Chi, Tatsunori Hashimoto, Oriol Vinyals, Percy Liang, Jeff Dean, and William Fedus. 2022.
\newblock \href {https://openreview.net/forum?id=yzkSU5zdwD} {Emergent abilities of large language models}.
\newblock \emph{Transactions on Machine Learning Research}.
\newblock Survey Certification.

\bibitem[{Wei et~al.(2018)Wei, Le, Dai, and Li}]{wei-etal-2018-airdialogue}
Wei Wei, Quoc Le, Andrew Dai, and Jia Li. 2018.
\newblock \href {https://doi.org/10.18653/v1/D18-1419} {{A}ir{D}ialogue: An environment for goal-oriented dialogue research}.
\newblock In \emph{Proceedings of the 2018 Conference on Empirical Methods in Natural Language Processing}, pages 3844--3854, Brussels, Belgium. Association for Computational Linguistics.

\bibitem[{Weidinger et~al.(2021)Weidinger, Mellor, Rauh, Griffin, Uesato, Huang, Cheng, Glaese, Balle, Kasirzadeh, Kenton, Brown, Hawkins, Stepleton, Biles, Birhane, Haas, Rimell, Hendricks, Isaac, Legassick, Irving, and Gabriel}]{weidinger2021ethical}
Laura Weidinger, John Mellor, Maribeth Rauh, Conor Griffin, Jonathan Uesato, Po-Sen Huang, Myra Cheng, Mia Glaese, Borja Balle, Atoosa Kasirzadeh, Zac Kenton, Sasha Brown, Will Hawkins, Tom Stepleton, Courtney Biles, Abeba Birhane, Julia Haas, Laura Rimell, Lisa~Anne Hendricks, William Isaac, Sean Legassick, Geoffrey Irving, and Iason Gabriel. 2021.
\newblock \href {https://arxiv.org/abs/2112.04359} {Ethical and social risks of harm from language models}.
\newblock \emph{Preprint}, arXiv:2112.04359.

\bibitem[{Weizenbaum(1972)}]{eliza}
Joseph Weizenbaum. 1972.
\newblock \href {https://doi.org/10.1126/science.176.4035.609} {On the impact of the computer on society}.
\newblock \emph{Science}, 176(4035):609--614.

\bibitem[{Welbl et~al.(2021)Welbl, Glaese, Uesato, Dathathri, Mellor, Hendricks, Anderson, Kohli, Coppin, and Huang}]{welbl-etal-2021-challenges-detoxifying}
Johannes Welbl, Amelia Glaese, Jonathan Uesato, Sumanth Dathathri, John Mellor, Lisa~Anne Hendricks, Kirsty Anderson, Pushmeet Kohli, Ben Coppin, and Po-Sen Huang. 2021.
\newblock \href {https://doi.org/10.18653/v1/2021.findings-emnlp.210} {Challenges in detoxifying language models}.
\newblock In \emph{Findings of the Association for Computational Linguistics: EMNLP 2021}, pages 2447--2469, Punta Cana, Dominican Republic. Association for Computational Linguistics.

\bibitem[{Xu et~al.(2023)Xu, Ju, Lane, Komeili, Smith, Ung, Behrooz, Ngan, Moritz, Sukhbaatar, Boureau, Weston, and Shuster}]{xu2023improving}
Jing Xu, Da~Ju, Joshua Lane, Mojtaba Komeili, Eric~Michael Smith, Megan Ung, Morteza Behrooz, William Ngan, Rashel Moritz, Sainbayar Sukhbaatar, Y-Lan Boureau, Jason Weston, and Kurt Shuster. 2023.
\newblock \href {https://arxiv.org/abs/2306.04707} {Improving open language models by learning from organic interactions}.
\newblock \emph{Preprint}, arXiv:2306.04707.

\bibitem[{Xu et~al.(2021)Xu, Ju, Li, Boureau, Weston, and Dinan}]{xu2021recipes}
Jing Xu, Da~Ju, Margaret Li, Y-Lan Boureau, Jason Weston, and Emily Dinan. 2021.
\newblock \href {https://arxiv.org/abs/2010.07079} {Recipes for safety in open-domain chatbots}.
\newblock \emph{Preprint}, arXiv:2010.07079.

\bibitem[{Yanai and Lercher(2020)}]{yanai2020thetwo}
Itai Yanai and Martin Lercher. 2020.
\newblock \href {https://doi.org/10.1186/s13059-020-02057-5} {The two languages of science}.
\newblock \emph{Genome Biology}, 21.

\bibitem[{Zang et~al.(2020)Zang, Rastogi, Sunkara, Gupta, Zhang, and Chen}]{zang-etal-2020-multiwoz}
Xiaoxue Zang, Abhinav Rastogi, Srinivas Sunkara, Raghav Gupta, Jianguo Zhang, and Jindong Chen. 2020.
\newblock \href {https://doi.org/10.18653/v1/2020.nlp4convai-1.13} {{M}ulti{WOZ} 2.2 : A dialogue dataset with additional annotation corrections and state tracking baselines}.
\newblock In \emph{Proceedings of the 2nd Workshop on Natural Language Processing for Conversational AI}, pages 109--117, Online. Association for Computational Linguistics.

\bibitem[{Zhang et~al.(2023)Zhang, Qian, Liu, Heinecke, Meng, Liu, Yu, Savarese, and Xiong}]{zhang2023dialogstudio}
Jianguo Zhang, Kun Qian, Zhiwei Liu, Shelby Heinecke, Rui Meng, Ye~Liu, Zhou Yu, Silvio Savarese, and Caiming Xiong. 2023.
\newblock Dialogstudio: Towards richest and most diverse unified dataset collection for conversational ai.
\newblock \emph{arXiv preprint arXiv:2307.10172}.

\bibitem[{Zhang et~al.(2018)Zhang, Dinan, Urbanek, Szlam, Kiela, and Weston}]{zhang-etal-2018-personalizing}
Saizheng Zhang, Emily Dinan, Jack Urbanek, Arthur Szlam, Douwe Kiela, and Jason Weston. 2018.
\newblock \href {https://doi.org/10.18653/v1/P18-1205} {Personalizing dialogue agents: {I} have a dog, do you have pets too?}
\newblock In \emph{Proceedings of the 56th Annual Meeting of the Association for Computational Linguistics (Volume 1: Long Papers)}, pages 2204--2213, Melbourne, Australia. Association for Computational Linguistics.

\bibitem[{Zheng et~al.(2023)Zheng, Chiang, Sheng, Zhuang, Wu, Zhuang, Lin, Li, Li, Xing, Zhang, Gonzalez, and Stoica}]{zheng2023judging}
Lianmin Zheng, Wei-Lin Chiang, Ying Sheng, Siyuan Zhuang, Zhanghao Wu, Yonghao Zhuang, Zi~Lin, Zhuohan Li, Dacheng Li, Eric Xing, Hao Zhang, Joseph~E. Gonzalez, and Ion Stoica. 2023.
\newblock \href {https://openreview.net/forum?id=uccHPGDlao} {Judging {LLM}-as-a-judge with {MT}-bench and chatbot arena}.
\newblock In \emph{Thirty-seventh Conference on Neural Information Processing Systems Datasets and Benchmarks Track}.

\bibitem[{Zhong and Ma(2022)}]{zhong2022effects}
Runting Zhong and Mengyao Ma. 2022.
\newblock \href {https://doi.org/10.1186/s12877-022-03428-2} {Effects of communication style, anthropomorphic setting and individual differences on older adults using voice assistants in a health context}.
\newblock \emph{BMC Geriatrics}, 22:751.

\end{thebibliography}
